\documentclass[9pt,twocolumn,twoside]{opticajnl}
\journal{science_robotics_arxiv}
\setboolean{shortarticle}{false}
\setboolean{displaycopyright}{false}

\def\scititle{Agile perceptive multi-skill locomotion\\
for quadrupedal robots in the wild}
\title{\scititle}

\author[1,2$\dagger$]{Jun-Gill Kang}
\author[2$\dagger$]{Jaehyun Park}
\author[2]{Tae-Gyu Song}
\author[2,3]{Joon-Ha Kim}
\author[4,5$\ast\ddagger$]{Seungwoo Hong}
\author[2$\ast\ddagger$]{Hae-Won Park}

\affil[1]{Agency for Defense Development, Daejeon 34186, Republic of Korea}
\affil[2]{Department of Mechanical Engineering, Korea Advanced Institute of Science and Technology, Daejeon 34141, Republic of Korea}
\affil[3]{DIDEN Robotics, Seoul 04799, Republic of Korea}
\affil[4]{School of Mechanical Engineering, Korea University, Seoul 02841, Republic of Korea}
\affil[5]{School of Smart Mobility, Korea University, Seoul 02841, Republic of Korea}
\affil[*]{Corresponding author: haewonpark@kaist.ac.kr; seungwoohong@korea.ac.kr}
\affil[$^\dagger$]{These authors contributed equally to this work.}
\affil[$^\ddagger$]{These authors contributed equally to this work.}
\dates{This is the accepted version of Science Robotics Vol. 11, Issue 116, adz7397 (2026)\newline DOI: 10.1126/scirobotics.adz7397}

\newcommand{\FigureOne}{%
\begin{figure*}[hbt!]
\centering
\includegraphics[width=\textwidth]{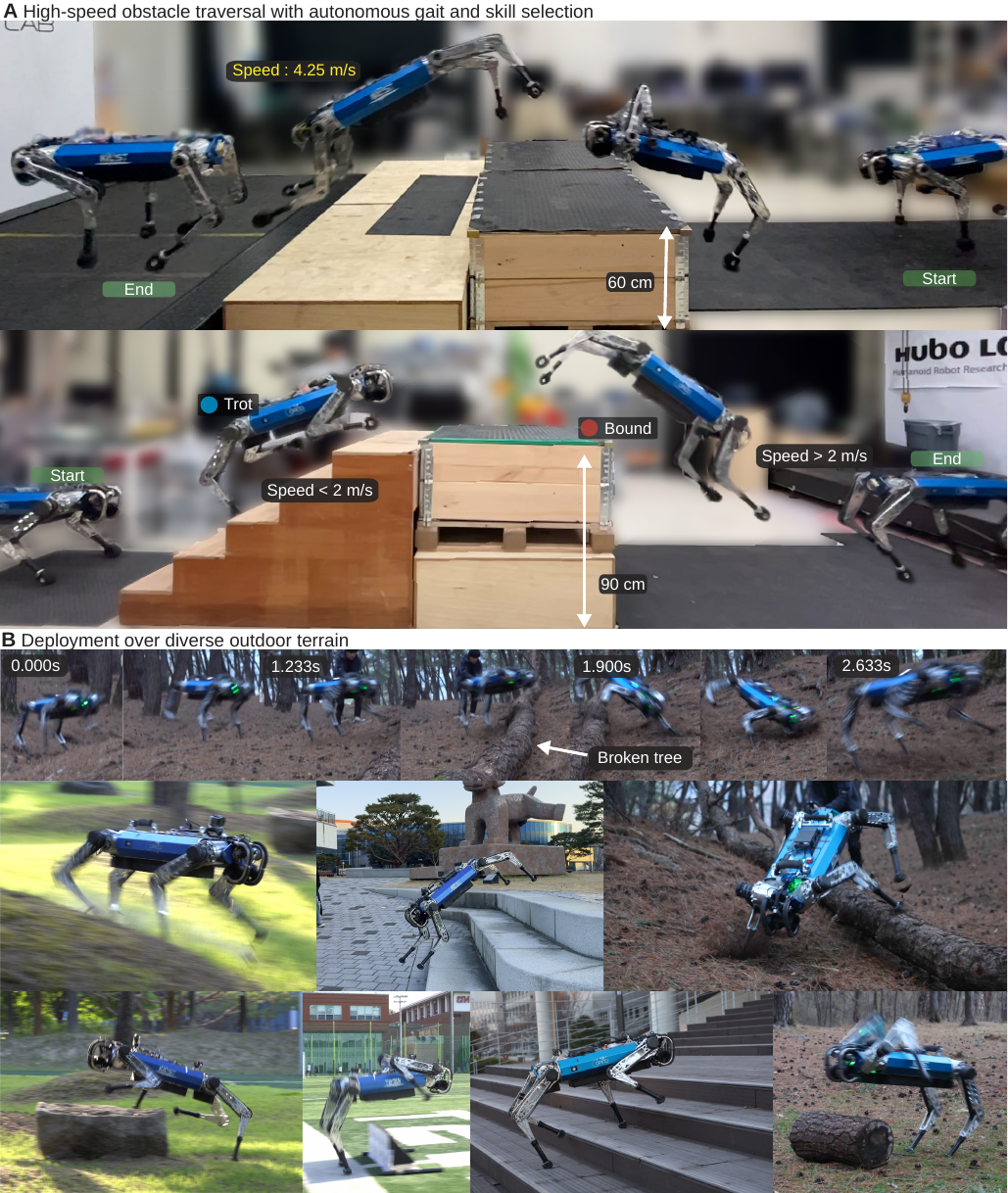}
\caption{\textbf{Traversing challenging terrains.} (\textbf{A}) Our controller enables fast perceptive gait and skill selection to enable high-speed obstacle traversal. (\textbf{B}) The same controller was used to deploy over various outdoor terrains.}
\label{fig:fig1}
\end{figure*}
}

\newcommand{\FigureTwo}{%
\begin{figure*}[hbt!]
\centering
\includegraphics[width=1.0\textwidth]{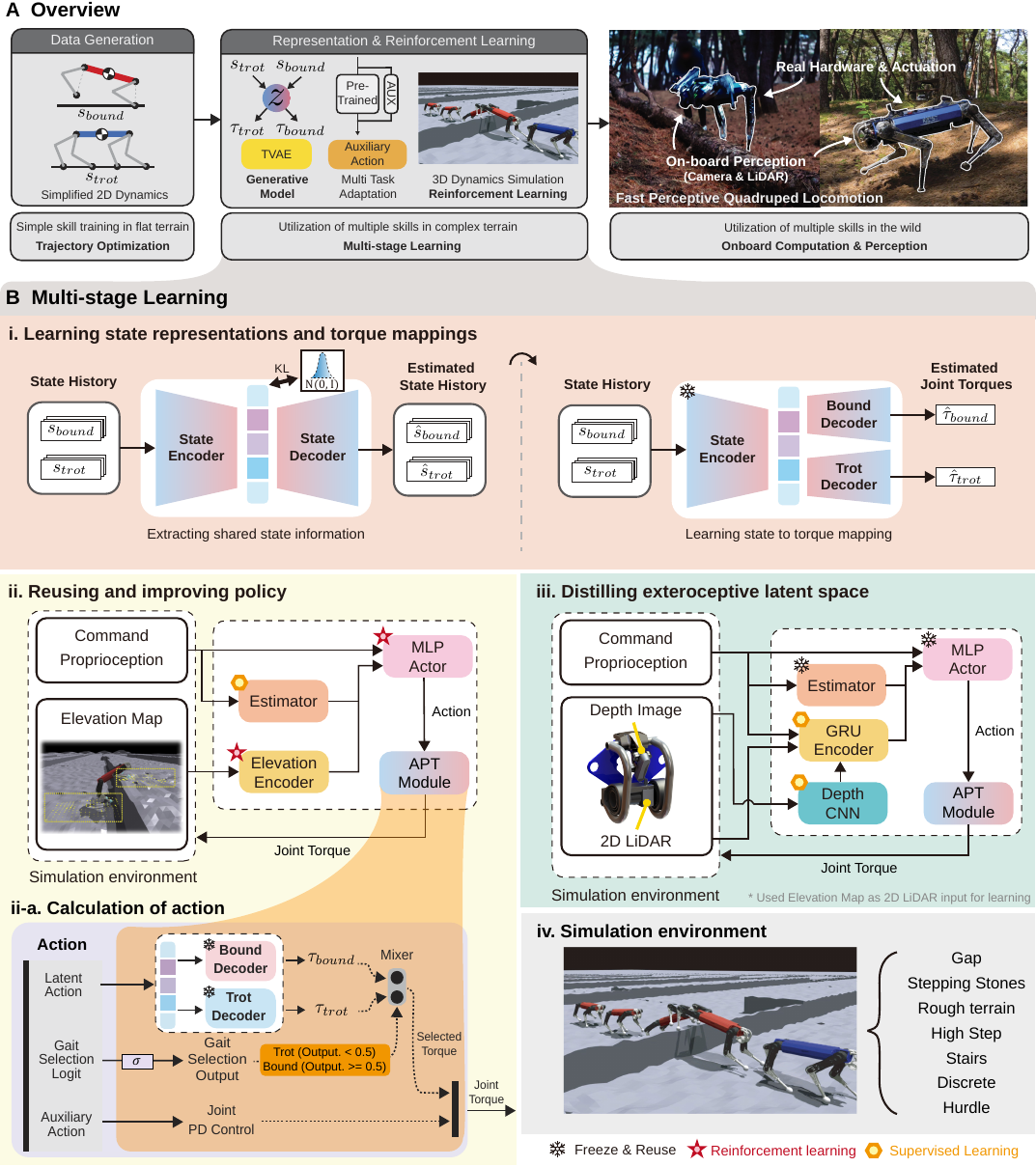}
\caption{\textbf{Training pipeline for APT-RL.} 
(\textbf{A}) Our approach consists of motion dataset generation, representation learning, and reinforcement learning. Using trajectory optimization with simplified dynamics, we generate large-scale motion data to train diverse locomotion skills. These serve as action priors for efficient downstream learning and enable autonomous locomotion in complex terrains.  (\textbf{B}) (\textbf{i}) Learning a shared latent state representation and gait-specific torque mappings using a TVAE.
(\textbf{ii}) Training policy by reusing the learned decoder via latent actions and adapting with auxiliary actions.
(\textbf{ii-a}) Calculation of joint torque using the APT Module.
(\textbf{iii}) Distilling exteroceptive latent space using CNN-GRU encoder.
(\textbf{iv}) Simulation environment.}
\label{fig:enc_dec}
\end{figure*}
}

\newcommand{\FigureThree}{%
\begin{figure*}[hbt!]
\centering
\includegraphics[width=0.95\textwidth]{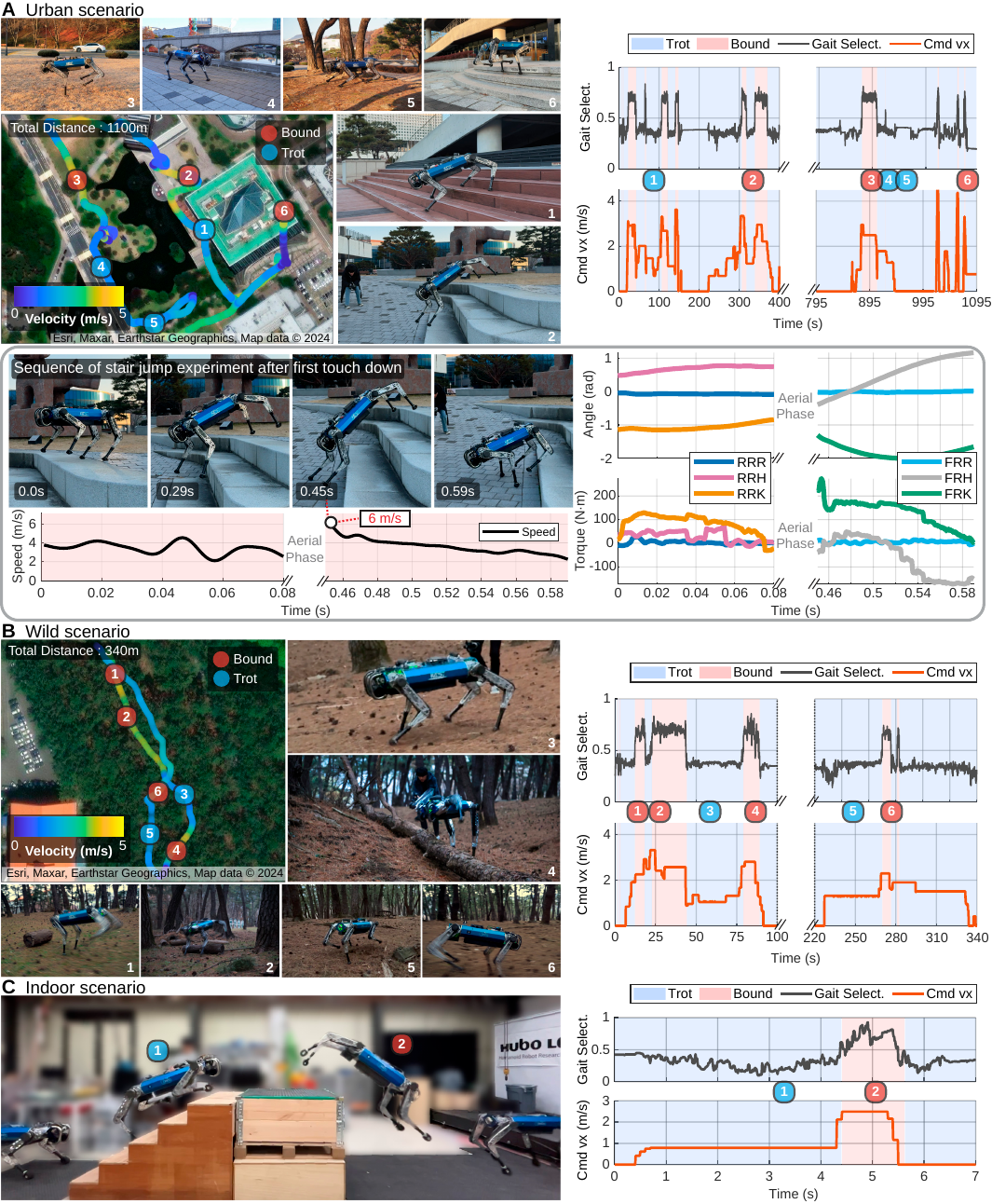}
\caption{\textbf{Agile perceptive locomotion with adaptive skills in various scenarios.} (\textbf{A}) Urban scenario, (\textbf{B}) Wild scenario, and (\textbf{C}) Indoor scenario. Cmd vx is the commanded forward velocity, which was visualized through color changes along the trajectory. Shaded blue regions indicate periods when the trotting gait was selected, whereas red indicates bounding. Gait Select. denotes the gait selection output, where values less than 0.5 correspond to trotting, and values greater than or equal to 0.5 correspond to bounding. To estimate the speed in the stair jump experiment, we computed the average velocity of the body center relative to the contact feet, followed by low-pass filtering at 60~Hz.}
\label{fig:various_scenarios}
\end{figure*}
}

\newcommand{\FigureFour}{%
\begin{figure*}[hbt!]
\centering
\includegraphics[width=0.75\textwidth]{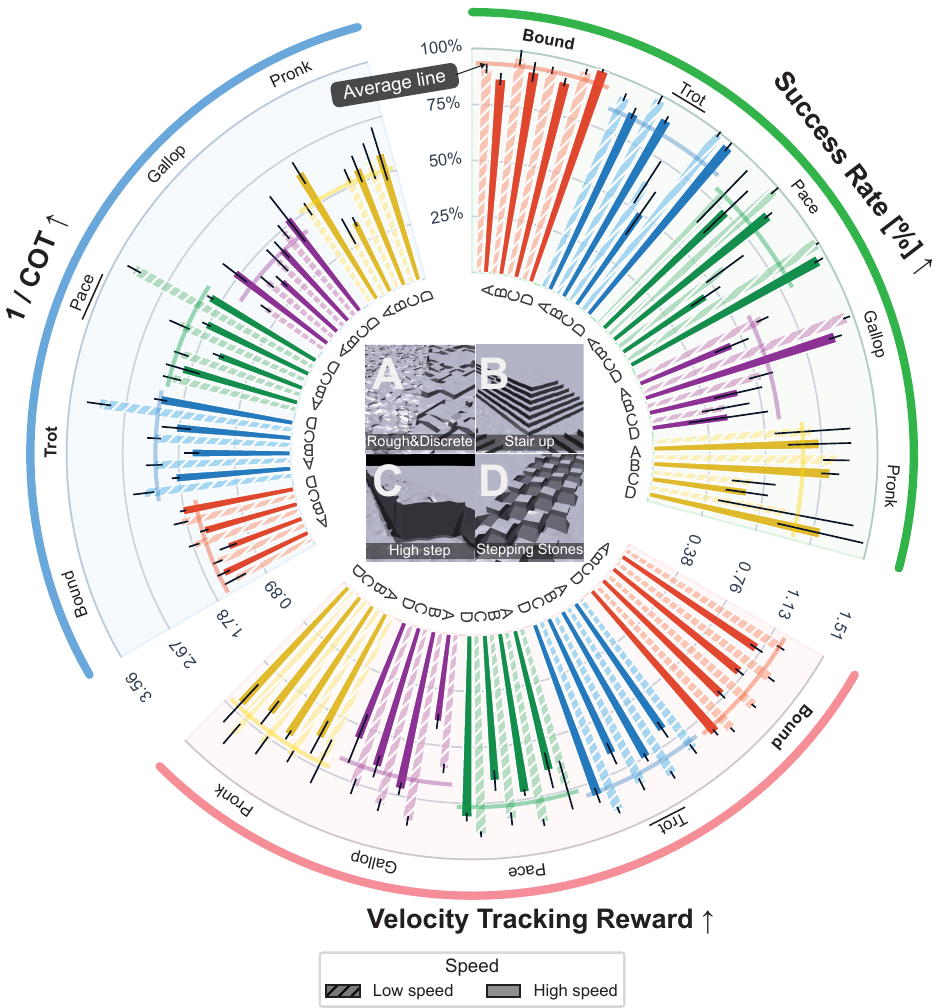}
\caption{\textbf{Effectiveness of gaits under different terrains.} Performance comparison of five gait decoders, trot, bound, pace, gallop, and pronk, across multiple terrain types. For the stair terrain, we used a lower command velocity range (train: -1 to 2m/s, test: 1 to 2m/s), whereas the other terrains were tested under a higher command velocity range (train: -1 to 7m/s, test: 1 to 7m/s). Each policy was trained with three random seeds and evaluated three times per seed, resulting in a total of nine trials with 300 agents per experiment, each receiving randomized velocity commands with randomized terrain difficulties. The average line in each sector represents the mean performance of each gait. The thick line segment on each bar denotes the standard deviation across repeated evaluations. The bold label marks the gait with the highest average score, and the underlined label marks the gait with the second-highest average score.}
\label{fig:multi_gait_terrain_evaluation}
\end{figure*}
}

\newcommand{\FigureFive}{%
\begin{figure*}[hbt!]
\centering
\includegraphics[width=1.0\textwidth]{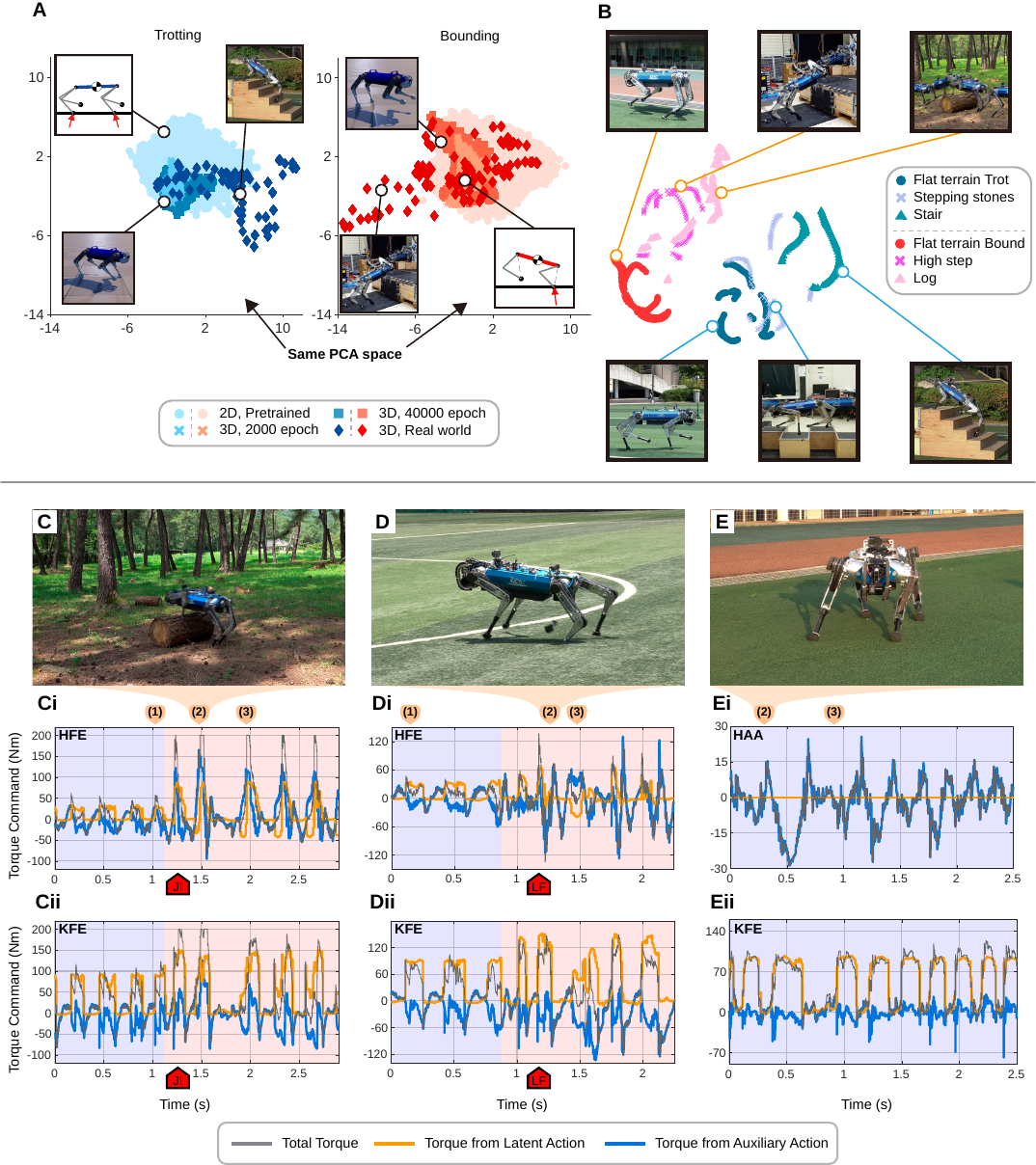}
\caption{\textbf{Latent action embeddings and auxiliary torque analysis.} 
(\textbf{A}) PCA embeddings of latent actions for trotting and bounding across pretraining, RL, and real-world deployment.  
(\textbf{B}) t-SNE embeddings of real-world latent actions on flat, stepping stones, stairs, high-step, and log terrains.  
(\textbf{C}--\textbf{E}) Representative real-world behaviors (log jump, leg fracture during walking, in-place rotation) together with their corresponding joint torque decompositions (\textbf{C(i)}--\textbf{E(ii)}), showing latent-action torque, auxiliary torque, and total torque.  
Blue and red shading indicate trotting and bounding selections; JI and LF denote the jump initiation and leg-fracture moments.}
\label{fig:dataset}
\end{figure*}
}

\newcommand{\FigureSix}{%
\begin{figure*}[hbt!]
\centering
\includegraphics[width=1.0\textwidth]{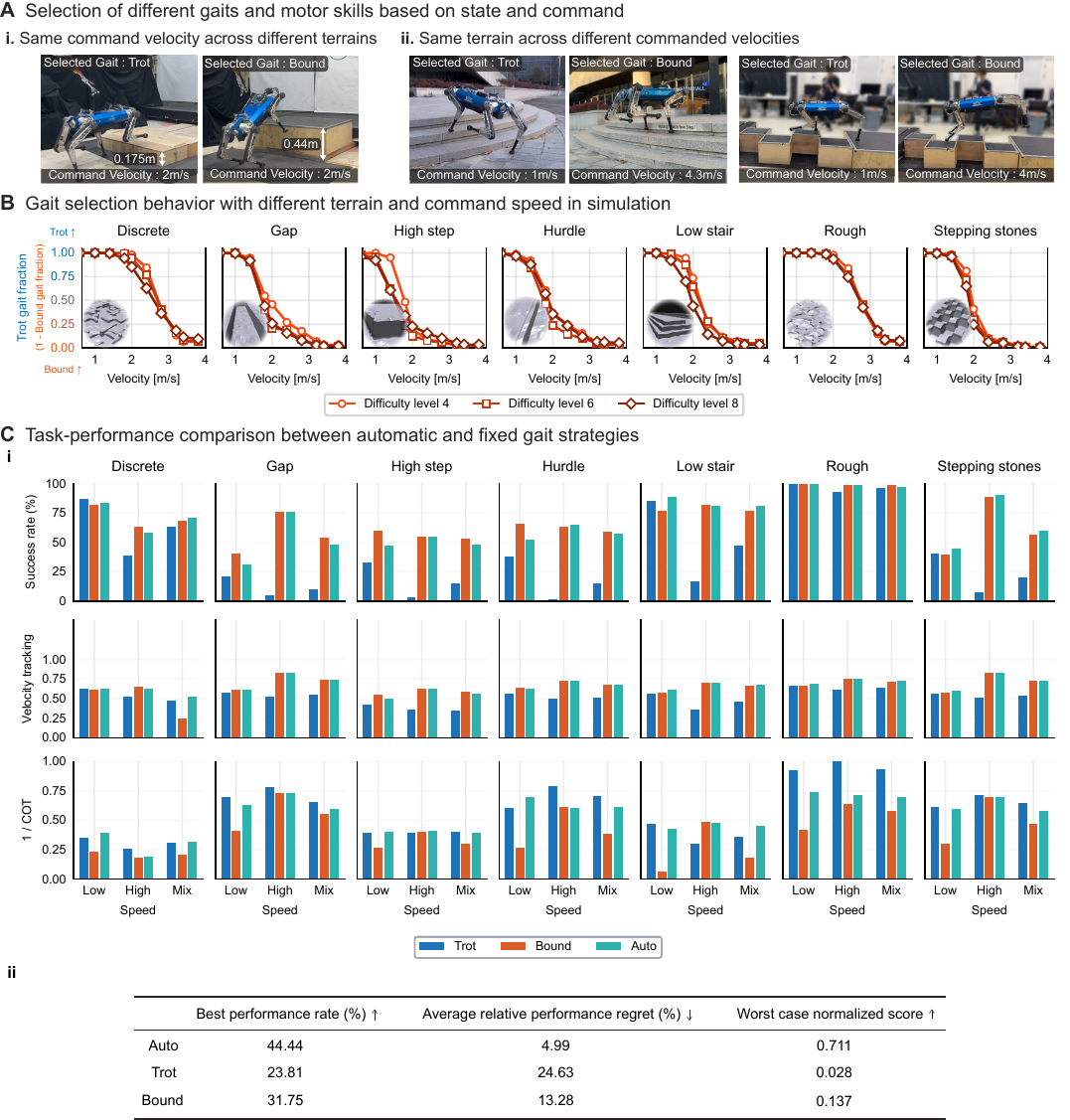}
\caption{\textbf{Autonomous gaits and motor skills selection.} 
(\textbf{A}) Our robot uses different gaits and motor skills to traverse obstacles, depending on its state and command.
(\textbf{B}) Gait fractions over continuous command velocities for various terrains and difficulty levels.
(\textbf{C}) Task-performance metrics for automatic and fixed gait strategies. 
(i) Success rate, velocity tracking, and 1/COT across three command-speed groups 
(Low: 0--3~m/s, High: 3--6~m/s, Mix: 0--6~m/s) and seven terrains. 
(ii) Aggregate comparison across all evaluation cases, showing for each controller the best-performance rate, the average relative performance regret, and the worst-case normalized score.}
\label{fig:trot_bound}
\end{figure*}
}

\newcommand{\FigureSeven}{%
\begin{figure*}[hbt!]
\centering
\includegraphics[width=0.85\textwidth]{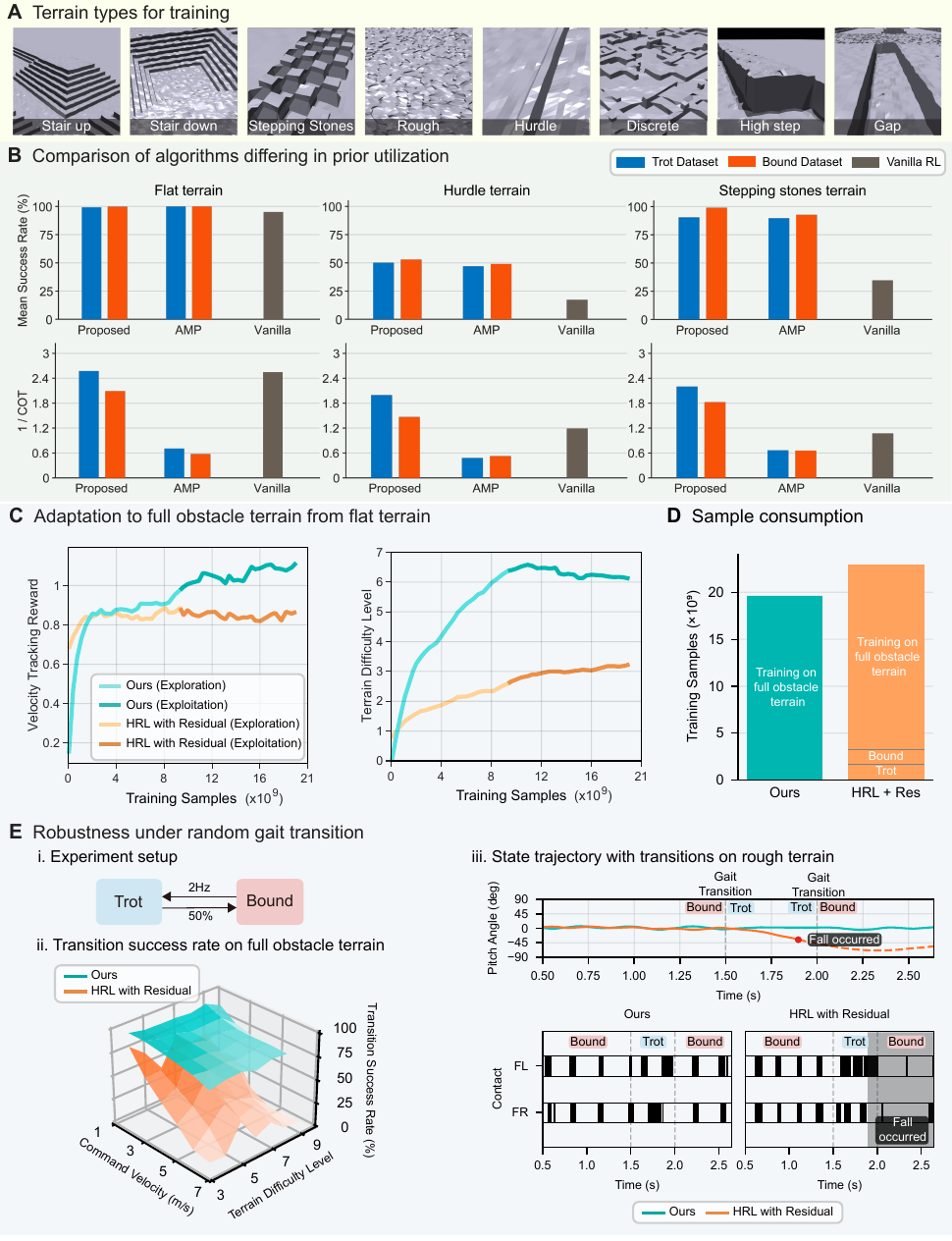}
\caption{\textbf{Evaluation across training strategies and baselines.} (\textbf{A}) Training terrains used across all experiments. (\textbf{B}) Comparison between our method, AMP, and Vanilla RL. (\textbf{C}) Comparison of training curves on full obstacle terrain between our method and HRL with a residual. (\textbf{D}) Comparison of sample consumption between our method and HRL with a residual baseline in full obstacle terrain. (\textbf{E}) Evaluation of dynamic gait transitions through random gait switching experiments. (i) Experimental setup for random gait transitions. (ii) Transition success rate with respect to terrain difficulty and command velocity. (iii) Base pitch angle trajectory and gait pattern with transitions on rough terrain.}
\label{fig:multi_skill_evaluation}
\end{figure*}
}

\newcommand{\FigureEight}{%
\begin{figure*}[hbt!]
\centering
\includegraphics[width=\linewidth]{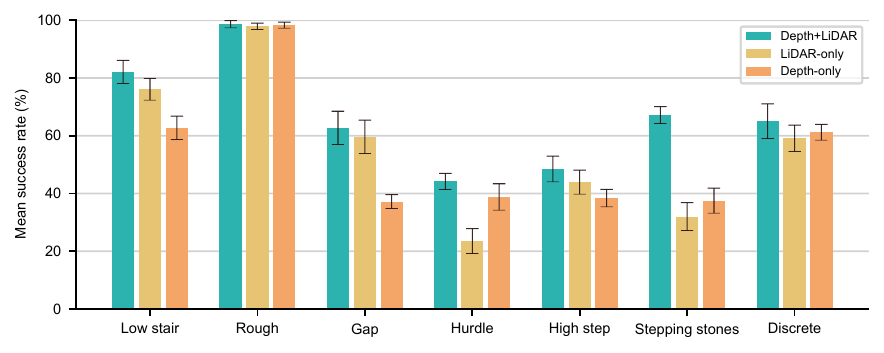}
\caption{\textbf{Sensor Ablation Study.} Success rate comparison of depth and LiDAR, LiDAR-only, and depth-only policies across diverse terrains. For each experiment, 100 agents were evaluated per trial over 10 trials, and the error bars represent the standard deviation.} 
\label{Fig:sensor_ablation}
\end{figure*}
}

\newcommand{\MovieOneFigure}{%
\begin{figure*}[t]
\centering
\includegraphics[width=0.75\textwidth]{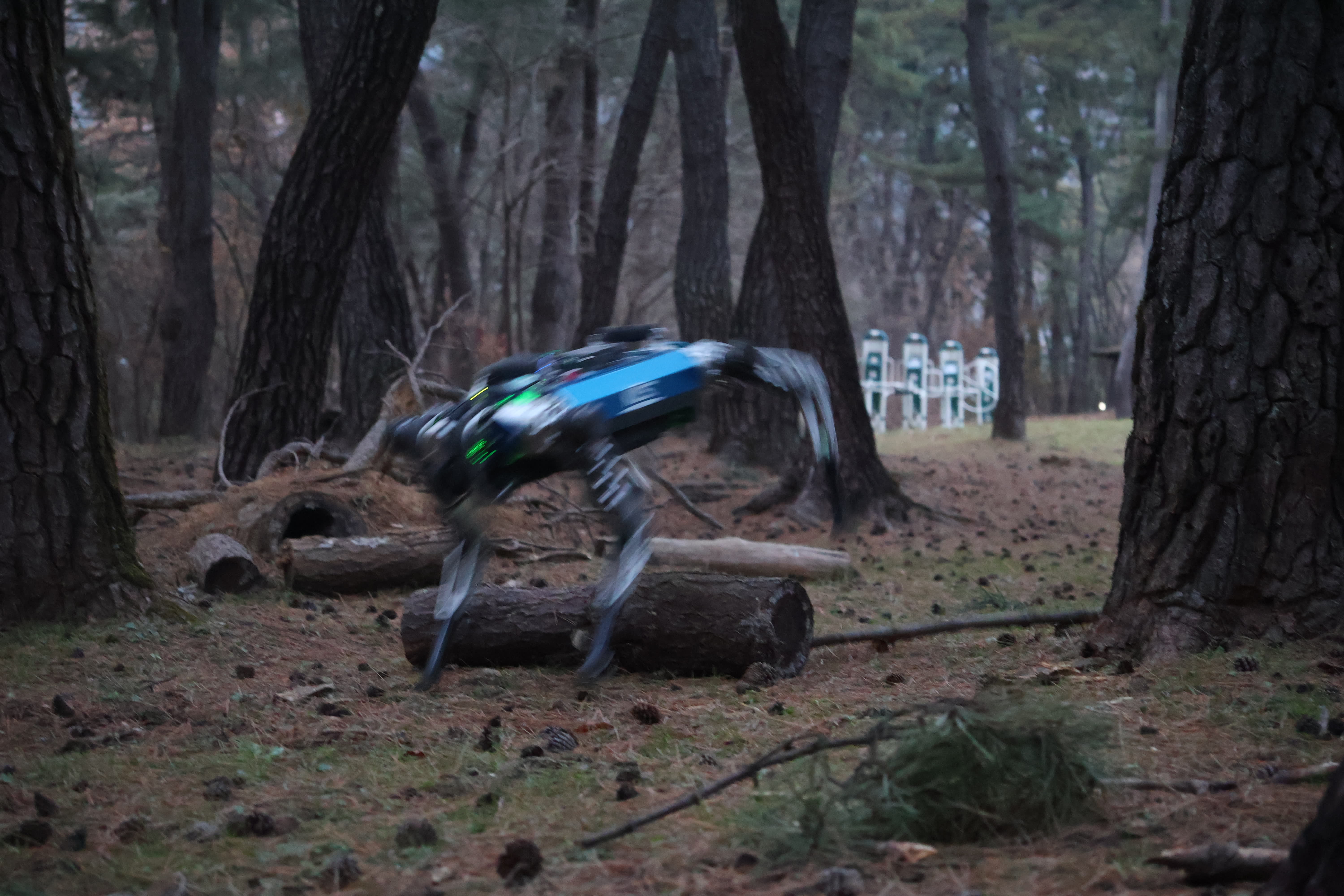}
{\captionsetup{labelformat=empty}
\caption{\textbf{Movie 1.} Summary of the results and the methods. The video clips at 0:19--0:25, 0:25--0:28, and 0:28--0:33 were adapted from \cite{youtube_traildogs}, \cite{youtube_agility}, and \cite{youtube_dogjump}, respectively, all licensed under a Creative Commons Attribution License (CC BY).}}
\label{fig:thumbnail}
\end{figure*}
}

\begin{abstract}
Enabling quadrupedal robots to traverse complex terrains---from rugged outdoor environments to urban landscapes---requires seamless integration of multiple motor skills, smooth transitions between gaits, and high-speed perceptive locomotion using only onboard sensors.
We present APT-RL (Action Pretrained Transformer-based Reinforcement Learning), a unified framework that enables multi-skill locomotion to achieve high-speed traversal in complex environments through autonomous skill transitions utilizing only onboard perception and computation.
Our approach generates large-scale, feature-rich 2D motion datasets through trajectory optimization with simplified dynamics. These datasets enable training of diverse, reusable locomotion skills that transfer effectively to a real quadruped robot operating on complex uneven terrains. The resulting high-quality skills serve as strong priors for efficient learning of complex downstream tasks and extend naturally to 3D environments, enabling smooth, high-speed multi-skill locomotion in deployed policy.
Real-world experiments demonstrate the framework's capabilities: the robot performs agile maneuvers through complex indoor obstacles and outdoor wild environments, including dynamic drop-down maneuvers that reach instantaneous peak speeds of up to 6 meters per second.  A single onboard policy enables robust traversal of diverse obstacles, including stairs, hurdles, stepping stones, gaps, and fallen branches, demonstrating the versatility and effectiveness of our approach.
\end{abstract}

\begin{document}

\maketitle

\section*{SUMMARY}
Quadruped robot autonomously selects diverse motor skills for perceptive high-speed locomotion on unstructured wild terrains
\section{Introduction}

Quadrupedal animals naturally adapt their gaits and motor skills to their physiology, environment, terrain, and locomotion speed \cite{hoyt1981gait, wimberly2021evolutionary}.
This adaptive behavior includes approaching obstacles through fast running and jumping, as well as combining various motor skills into precisely timed and coordinated sequences. Consequently, quadrupeds agilely traverse complex terrains by using multiple locomotion skills while maintaining stability and agility. Likewise, legged robots have the potential to reach otherwise inaccessible locations by dynamically and efficiently traversing complex terrains, including rubble, steps, and logs, at high speeds. Enabling such dynamic locomotion is crucial not only for navigating natural outdoor environments such as forests and mountainous regions, but also for advancing robotic performance in high-stakes applications such as search-and-rescue missions \cite{martin2020cadaver, fenton1992use, jones2004search}.

To realize this goal, legged robots must be equipped with several key components: first, a robust control architecture that supports multiple gaits and corresponding motor skills across a broad range of speeds; second, the ability to autonomously perform smooth, flexible, stable gait and skill transitions that adapt to hard-to-reach terrain conditions; and third, an efficient and practical learning framework that enables the acquisition and deployment of complex locomotion behaviors on real-world robotic platforms. These elements must be harmoniously integrated and adapted to diverse environments, including the precise coordination of multiple agile motor skills. Developing quadrupedal control systems capable of traversing diverse terrains has long been a central pursuit within the legged robotics community~\cite{ijspeert2008central, burden2024animals}, motivating extensive research in both model-based control and reinforcement learning (RL) approaches.

Model-based control methods, grounded in physical dynamics and optimization techniques, have demonstrated dynamic and versatile locomotion, from zero-moment-point control \cite{kalakrishnan2010fast}, static walking \cite{semini2011design}, and dynamic walking \cite{hutter2012starleth}, and running \cite{hyun2014high, park2015online, park2015variable, hutter2016anymal, park2017high} to more recent nonlinear model predictive control (MPC) \cite{hong2020real, ding2021representation, hong2022agile, shin2022design, grandia2023perceptive} and contact-implicit MPC \cite{kim2025contact}. 
However, although effective under modeled conditions, the difficulty of accurately modeling and estimating complex contact dynamics, friction, and environmental interactions limits the adaptability and robustness of model-based controllers.
In contrast, model-free RL methods handle model uncertainty without explicit dynamics modeling, enabling robust and agile control across blind flat terrain \cite{tan2018sim, iscen2018policies, hwangbo2019learning,shin2024reinforcement}, blind rough terrain \cite{lee2020learning, ji2022concurrent, choi2023learning, nahrendra2023dreamwaq, chen2023learning, kim2024learning, margolis2024rapid, kim2024not}, adaptive \cite{yu2017preparing, kumar2021rma}, perceptive \cite{rudin2022learning, miki2022learning, miki2024learning, luo2024pie, made2024obstacle}, and biologically inspired \cite{bellegarda2022cpg} locomotion settings.
Nonetheless, such methods frequently encounter challenges such as high sample complexity and require carefully tuned reward functions to produce realistic and dynamic behaviors. 
Furthermore, training a single end-to-end policy to accommodate multiple gaits and motor skills can degrade performance relative to task-specific policies tailored to particular environments. Consequently, many existing works focus on relatively flat terrains to achieve high-speed locomotion \cite{jin2022high, margolis2024rapid, shin2024reinforcement}, whereas controllers designed for obstacle navigation are typically limited to moderate speeds (approximately 1~m/s) \cite{zhuang2023robot, cheng2023extreme, hoeller2024anymal}.
As a result, these methods are unsuitable for traversing more complex terrains, as they lack the essential capability to execute diverse gaits and motor skills across a broad range of speeds, particularly at high speeds.
A recent study combining RL with a random sampling planner demonstrated impressive locomotion over diverse terrains in controlled indoor environments \cite{kim2025high}, but its limited gait diversity and reliance on external state estimation, such as motion capture instead of onboard sensors, make it unsuitable for outdoor and unstructured environments.

Many approaches leverage prior guidance, such as expert demonstrations from animal motion or optimized trajectories, to encourage robust behaviors and reduce reward-tuning effort. 
One prominent example is the use of quadrupedal animal motion-capture data in Adversarial Motion Prior (AMP) \cite{peng2021amp}, which enables robots to reproduce animal-like locomotion without manually designed complex objective functions \cite{escontrela2022adversarial, han2024lifelike}. 
However, collecting large-scale animal motion datasets across diverse terrains is challenging and costly. 
As an alternative, trajectory optimization (TO) with an appropriate robot dynamics model can generate physically feasible motion data, synthesizing high-quality motions and corresponding torque commands without relying on animal data.
Such approaches provide systematic, physically grounded motion references that can guide policy learning and generate scalable, user-preferred, and task-specific motions \cite{jin2022high, brakel2022learning, fuchioka2023opt, kang2023rl+, wu2023learningamp, jenelten2024dtc}. 
Nevertheless, most of these methods \cite{peng2021amp, escontrela2022adversarial, jin2022high, brakel2022learning, fuchioka2023opt} focus on tracking predefined reference data, which limits adaptability to new tasks or substantial environmental variations. 
Similarly, \cite{wu2023learningamp} uses only TO state trajectories as motion priors, requiring an additional layer to reconstruct corresponding actions and retraining when encountering new terrain. Other methods \cite{kang2023rl+, jenelten2024dtc} solve TO online during RL training to provide context-aware policy references; however, online TO remains computationally expensive.

Another line of work pretrains motion skills for specific environments and later selects or composes them to improve generalization.
Hierarchical Reinforcement Learning (HRL) \cite{barto2003recent} and Mixture-of-Experts (MoE) \cite{shazeer2017outrageously} architectures have been used to achieve adaptive locomotion by leveraging pretrained skills. However, they remain limited in achieving smooth skill transitions under changing terrain conditions and in reusing skills across previously unseen environments. 
For instance, ANYmal Parkour \cite{hoeller2024anymal} employs an HRL framework to select obstacle-specific skills from visual perception, but deployment is limited to environments containing only the obstacles used when training the low-level skills.
Similarly, Multi-Expert Learning Architecture (MELA) \cite{yang2020multi} adopts an MoE architecture combining locomotion modes such as trotting and fall recovery, allowing transitions between them. Nonetheless, it lacks diversity in motion skills and does not fully address how skill composition can be integrated with perception for traversing complex and challenging environments.

To enrich motion diversity and naturally encode transitions, researchers have also drawn inspiration from biological systems. For example, Central Pattern Generators (CPGs) \cite{shafiee2024viability, bellegarda2025allgaits, owaki2017quadruped} incorporate prior knowledge of animal sensorimotor coordination and enable smooth transitions across different gaits due to their effectiveness in modeling periodic and transitional dynamics. Notably, \cite{shafiee2024viability} demonstrates seamless gait transitions in quadrupeds, providing the important insight that energy efficiency is not the sole factor driving gait switching. Similarly, other researchers aim to achieve smooth motion transitions and increase training efficiency through the reuse of prior motor skills, representing diverse locomotion behaviors within a shared latent space in which a controller is trained to operate \cite{hasenclever2020comic, liu2022motor, peng2022ase, luo2023universal}.

In robotics, this paradigm has been adopted to pretrain locomotion skills in a general latent space that can later be repurposed for downstream tasks, including ball dribbling \cite{bohez2022imitate}, adaptive gait generation \cite{mitchell2023vae, wu2023learninggait, mitchellgaitor}, and parkour-like motions \cite{han2024lifelike}.
Although these methods demonstrate broad applicability by leveraging diverse motion embeddings, they either require an additional low-level controller or the learning of an action policy to track or match the reference motion data.
For instance, \cite{bohez2022imitate, wu2023learninggait, han2024lifelike} rely on state-only motion data without corresponding torques, necessitating an additional RL stage to recover control commands, whereas VAE-Loco \cite{mitchell2023vae} and Gaitor \cite{mitchellgaitor} depend on whole-body controllers (WBCs) during deployment to track the planned base motion and contact schedules. This decoupled design limits scalability to high-speed locomotion, where WBCs often struggle to handle fast contact dynamics and unstructured terrain interactions.
Despite their promise, these methods have not yet been demonstrated for perceptive locomotion in real-world, high-speed, or highly unstructured environments.

\FigureOne

We propose Action Pretrained Transformer-based Reinforcement Learning (APT-RL), a unified control architecture designed to enable robust locomotion across a wide range of speeds with diverse gaits and motor skills (Fig.~\ref{fig:fig1}A and B), facilitate flexible yet stable skill transitions for traversing challenging terrains (Fig.~\ref{fig:fig1}A), and provide efficient learning. 
The robot acquires locomotion capabilities through a multi-stage progression, gradually learning from simple to complex motor behaviors. In the early stages, it learns basic gait patterns, such as trot, bound, through diverse motions on flat terrain, which are later reused and adapted to more complex environments.
Our key insight lies in generating and collecting large-scale, feature-rich 2D robot motion datasets using trajectory optimization (TO), while keeping the data generation process affordable and easily obtainable.
These datasets are used to learn diverse and reusable locomotion skills for quadruped robots operating in diverse extreme uneven terrain (Fig.~\ref{fig:enc_dec}A).
Specifically, we use Single Rigid Body Dynamics (SRBD), which captures salient full-body dynamics with substantially reduced model dimensionality and complexity. This simplification not only reduces computational overhead and enables fast data generation, such as producing 180,000 trajectories with a total duration of 15.5~hours in just 8~minutes, but also preserves motion quality suitable for downstream learning. 
Moreover, unlike existing methods \cite{bohez2022imitate, peng2022ase, luo2023universal, wu2023learningamp, wu2023learninggait, han2024lifelike} that require additional RL to match reference motion, our framework does not require a separate RL stage to train a reference-tracking policy. 
Specifically, our method employs a Transformer-based variational autoencoder (TVAE), trained on the TO-generated dataset, to learn a unified latent representation of
TO-derived knowledge and to acquire a reusable torque decoder through imitation learning,
without requiring any RL-based representation learning or online TO during training. 
This is enabled by trajectory optimization, which jointly generates motion trajectories and corresponding control inputs.
The resulting high-quality locomotion skills serve as strong priors, enabling efficient exploration and learning of complex downstream tasks. 
Furthermore, these skills can be seamlessly extended to 3D and leveraged for multi-skill high-speed perceptive locomotion with smooth transitions in the final deployed policy.

We validate the effectiveness of APT-RL through zero-shot sim-to-real transfer in challenging real-world environments. Our controller enables the KAIST HOUND \cite{shin2022design} robot to traverse unstructured natural environments, including campus pathways and forested regions, using only onboard LiDAR, a depth camera, and onboard computation. The robot performs dynamic maneuvers over obstacles such as high steps, stairs, hurdles, logs, fallen branches, gaps, and stepping stones.
Thanks to our well-structured learning phases and high-quality reusable motion priors, the final deployed policy autonomously acquires multi-skill and gait transition capabilities, exhibiting maneuver-level behaviors such as jumping over obstacles, high-step climbing, and drop-down landings, selecting appropriate gaits and motor skills to navigate challenging terrains at high speed.

KAIST HOUND performs agile maneuvers through complex indoor obstacles and outdoor wild environments, achieving an instantaneous peak speed of 4.25 m/s while overcoming a high step, which involves sequential combined jump-up and drop-down events. In addition, the robot reaches an instantaneous peak speed of 6 m/s while jumping down a three-step staircase, measured immediately before ground impact during the drop-down transition. The corresponding Froude numbers are 3.85 (at 4.25 m/s) and 7.69 (at 6 m/s), respectively, establishing a new benchmark for perceptive quadrupedal locomotion in challenging real-world conditions.

\section{Results}

Our APT-RL consisted of three integrated phases. The representation learning phase constructed a structured latent space from large-scale TO motion data (Fig.~\ref{fig:enc_dec}B-i). The reinforcement learning phase incorporated an auxiliary action while leveraging the decoder trained during representation learning (Fig.~\ref{fig:enc_dec}B-ii). The perceptual distillation phase enabled deployment with real-world sensory inputs (Fig.~\ref{fig:enc_dec}B-iii).

In the representation learning phase (Fig.~\ref{fig:enc_dec}B-i), we learned a structured latent space and a torque decoder to support diverse motion generation, smooth transitions, and efficient learning. We first generated large-scale torque-annotated motion data using 2D TO and used it to train a TVAE \cite{kang2023highly, petrovich2021action}, which formed the unified latent space. In this phase, reconstruction-based self-supervised learning captured continuous structure across speeds and gait characteristics. In a subsequent step, supervised learning trained a decoder that maps latent representations to torque commands.

The reinforcement learning phase (Fig.~\ref{fig:enc_dec}B-ii) learned a policy that selected latent actions from observations to adaptively handle diverse obstacle environments by leveraging a pretrained torque decoder in combination with an auxiliary action. This enabled gait and skill selection and smooth transitions. The policy employed an auxiliary structure (Fig.~\ref{fig:enc_dec}B-ii-a), where the auxiliary action complemented the torque generated by decoding a latent action. This overall architecture supported the generation of a wide range of speeds and gait patterns, as well as context-aware transitions and refinements.

To enable real-world deployment, we performed exteroceptive information distillation (Fig.~\ref{fig:enc_dec}B-iii). The student exteroceptive encoder, which took perceptual sensory inputs, was trained to mimic the output of a teacher encoder based on privileged height maps. The perceptual sensory inputs, consisting of depth camera images and 2D LiDAR data, ensured robust and responsive perception during fast and dynamic locomotion. During real-robot experiments, custom-made mechanical vibration absorbers were installed on the robot body to improve sensor stability under physical shocks and vibration, as detailed in the Supplementary Materials.

\FigureTwo

\subsection*{Fast perceptive quadruped locomotion in complex terrain}

We deployed the developed locomotion controller on our in-house quadrupedal robot, KAIST HOUND, across various unstructured wild environments, urban settings, and indoor obstacle courses, as shown in Fig.~\ref{fig:fig1}. We successfully deployed the robot both indoors and outdoors on a variety of terrain surfaces, including rough grass, debris-covered forest floors, asphalt surfaces, and obstacle-laden environments such as stairs, high steps, stepping stones, gaps, and unstructured rocky terrains. See Movie 1 for a video demonstration of the experiments.

\subsection*{Adaptive skill selection in urban, wild, and indoor environments}

To evaluate the effectiveness of the proposed controller's skill selection ability under real-world conditions, we conducted long-duration navigation experiments in two distinct outdoor environments: an urban campus and a wild forest trail (movie S1). In addition, we evaluated the controller in an indoor obstacle course with varying obstacle types.

\FigureThree

In the urban scenario, the robot successfully traversed a 1.1~km urban environment featuring challenging structured terrains such as multi-level stairs, grass patches, and inclined ramps (Fig.~\ref{fig:various_scenarios}A). Throughout the course, the robot adapted its gait in response to terrain characteristics, proprioceptive states, and command velocity. Notably, in Fig.~\ref{fig:various_scenarios}A-1, the robot employed a trotting gait to descend a staircase at a lower command speed (1~m/s), whereas in  Fig.~\ref{fig:various_scenarios}A-6, it used bounding to traverse stairs at higher command speeds (4.3~m/s). During stair-jumping-down experiments, the robot reached instantaneous peak speeds of up to 6~m/s while descending a three-step staircase, with each step measuring 30~cm in height and 58~cm in depth.

In the wild scenario, the robot completed a 0.34~km traversal of unstructured forest terrain, including fallen trees, exposed roots, uneven logs, and leaf-covered or slippery ground (Fig.~\ref{fig:various_scenarios}B). It demonstrated agile behavior by autonomously selecting appropriate gaits across various scenes (Fig.~\ref{fig:various_scenarios}B-1 to 6): bounding over elevated logs (Fig.~\ref{fig:various_scenarios}B-1, 2, and 4), switching to trotting in irregular regions (Fig.~\ref{fig:various_scenarios}B-3 and 5), and returning to bounding as the commanded velocity increased (Fig.~\ref{fig:various_scenarios}B-6). 

In the indoor obstacle course scenario (Fig.~\ref{fig:various_scenarios}C), the robot used a trotting gait to climb stairs and transitioned to bounding before jumping off a 90~cm vertical step. Specifically, at 4.41 seconds, it switched to bounding while modulating footstep timing prior to takeoff. In a separate indoor scenario shown in Fig.~\ref{fig:fig1}A, the robot used a bounding gait to leap over a 60~cm-high step, reaching an instantaneous peak speed of 4.25~m/s while traversing the obstacle.

Collectively, these results show that our policy autonomously selects both gaits and motor skills, conditioned on terrain geometry, proprioceptive state, and commanded velocity. Effectiveness of different gaits across various terrains, analysis of the policy's latent action and torque utilization, as well as the policy's adaptive gait selection behavior, are presented in subsequent sections.

\FigureFour

\subsection*{Effectiveness of gaits under different terrains}

In this section, we present the motivation for selecting trot and bound as the two primary torque-level gait primitives by analyzing the relationship between gait type, terrain characteristics, and agility. To support this selection, we evaluated a broader set of gait decoders under diverse terrain conditions, including rough~\&~discrete terrain, stairs, high steps, and stepping-stone environments.

We trained five gaits decoder, trot, bound, pace, gallop, and pronk, each pretrained using the same trajectory-optimization procedure and trained under identical conditions to ensure a fair comparison. As shown in the Fig. \ref{fig:multi_gait_terrain_evaluation}, we used the Success Rate, Velocity Tracking Reward, and the inverse of the Cost of Transport (1/COT) as the main evaluation metrics. Higher values indicate better performance for all metrics.

Overall, our findings indicate that trot and bound provided the most reliable and complementary performance across terrains. Trot generally demonstrated strong stability and a low Cost of Transport. In contrast, bound consistently achieved higher success rates in high-speed regimes and on terrains with large height variations, such as high-step obstacles.

Although additional gaits offered advantages in specific situations, their overall performance was less consistent. For example, pacing often exhibited competitive energy efficiency, but its success rate and velocity-tracking reward were generally lower than those of trot. Gallop and pronk occasionally achieved lower COT compared to bound, but they showed larger variance across trials and lower success rates, making them less suitable for general-purpose deployment.

Taken together, these findings support our selection of trot and bound as the primary gaits, as they offered the most complementary and consistently effective behaviors among those evaluated.

\subsection*{Generation of diverse 3D motion from 2D quadruped robot trajectory samples}

To develop a controller that generates diverse motor skills across gaits and smooth transitions between them, we employed a representation learning approach by training a decoder on data produced via 2D TO only on flat terrain. We selected 2D TO for its ability to capture salient physical dynamics relevant to locomotion at reduced computational cost, enabling efficient extraction of reusable features. To examine how the policy leveraged the learned decoder to produce both familiar and novel motions, we analyzed the latent actions generated by the pretrained encoder and the policy. This analysis provides insight into how the policy orchestrates learned representations to achieve versatile and adaptive control.

Figure~\ref{fig:dataset}A presents a principal component analysis (PCA) visualization of latent actions across three stages of the pipeline, including the pretrained dataset, policy training, and real-world robot execution. We assume that the latent space follows a standard Gaussian distribution; a detailed explanation of our latent space modeling is provided in the Materials and Methods section. Principal components were computed using combined latent actions from trotting and bounding in the 2D TO dataset (Fig.~\ref{fig:dataset}A), and a transformation was applied to training and deployment data for direct comparison within the same space. In the early stages of training, latent actions generated by the policy were concentrated within a subset of this pretrained cluster and gradually expanded outward as learning progressed. During real-world deployment, the policy drew on latent actions overlapping with pretrained regions for familiar gaits, yet extended into novel areas when handling more challenging or unseen terrains.

Figure~\ref{fig:dataset}B illustrates a t-distributed stochastic neighbor embedding (t-SNE) visualization of latent actions generated by the robot across terrains using trot and bound gaits. Latent actions exhibited clear clustering by gait type, with additional sub-clusters within each group corresponding to specific motion variations. This structure reflects how the latent space naturally organizes according to behavior similarity. These findings indicate that, although the latent space was shaped solely by 2D TO data from flat terrain rather than by the final 3D behaviors, the policy learned to leverage its underlying structure through interaction during RL training.

\FigureFive

\subsection*{Robust policy adaptation with auxiliary action}

To evaluate how the policy combines the pretrained decoder with the auxiliary action, we analyzed the contributions of two torque sources: the torque produced by the pretrained decoder (decoder torque) and the torque computed from the auxiliary action using a low-level PD controller (auxiliary torque), through RL training. The decoder, trained on 2D TO data, acts as a structured torque generator conditioned on latent actions from the RL policy. However, because the 2D data lacks coverage of full 3D dynamics and terrain complexity, it alone is insufficient for robust adaptation.

To address this, the policy jointly learned an auxiliary action that refined the decoder's output, enabling adaptation to diverse terrain conditions. This auxiliary torque complemented the decoder torque, allowing the final policy to generate motions beyond the coverage of the pretrained decoder. Unlike traditional residual policies \cite{silver2018residual, johannink2019residual}, which refine a fixed base policy, our method jointly learned both latent and auxiliary actions, supporting integrated refinement.

We analyzed the composition of total torque from latent and auxiliary actions in three representative scenarios (Fig.~\ref{fig:dataset}C-E). In the log-jumping scenario (Fig.~\ref{fig:dataset}C), which represented an unseen scenario not present in the pretraining dataset, we analyzed the torque contributions from the rear-left leg's hip flexion-extension (HFE) and knee flexion-extension (KFE) motors (Fig.~\ref{fig:dataset}C-i, ii). In Fig.~\ref{fig:dataset}C-i, before crossing the log on flat terrain, the motion was primarily driven by decoder torque. As the jump began, the auxiliary torque became more prominent, enabling the generation of a jumping motion absent from the pretraining dataset.

In the leg-breakage scenario (Fig.~\ref{fig:dataset}D), which reflected an out-of-distribution event not encountered during training, we analyzed the same motors (Fig.~\ref{fig:dataset}D-i, ii). After breakage, the policy reduced the decoder torque of the HFE motor and increased the auxiliary torque, altering the leg trajectory and allowing the robot to maintain balance.

The pretraining dataset lacked torque data for the hip abduction-adduction (HAA) motor, resulting in no decoder torque for this joint. To examine how the policy compensated for this missing decoder torque, we analyzed torque contributions from the rear-left leg's HAA and KFE motors during in-place rotation (Fig.~\ref{fig:dataset}E-i, ii). Whereas the KFE motor was primarily driven by decoder torque, the HAA motor relied on auxiliary torque when decoder torque could not be generated, enabling yaw rotation, a behavior not present in the original dataset. These results demonstrate that our policy leveraged both pretrained torque patterns from the TO dataset and learned auxiliary adjustments, enabling robust adaptation and the generation of novel 3D motions beyond the original pretraining data. A video demonstrating the experiments is provided in movie S2.

\FigureSix

\subsection*{Autonomous gaits and motor skills selection}

To evaluate the proposed controller's gait and motor skill selection, we analyzed the policy across diverse terrains and command speeds. Inspired by animals' flexible gait selection and the detailed gait-terrain ablation study in Fig.~\ref{fig:multi_gait_terrain_evaluation}, we designed the policy to experience both trotting and bounding, along with their corresponding motor skills, and to select between them based on current observations.

Figure~\ref{fig:trot_bound}A shows real-world demonstrations in which the policy selected gaits based on the state, terrain, and commanded velocity. In Fig.~\ref{fig:trot_bound}A-i, with a fixed 2~m/s command, the policy chose trot for a lower obstacle (0.175~m) and bound for a higher obstacle (0.44~m). In Fig.~\ref{fig:trot_bound}A-ii, on the same terrain, the policy used trot at 1~m/s and switched to bound above 4~m/s (movie S3).

For further examination, we conducted simulation experiments across terrains and difficulty levels, with command speeds ranging from 0.5 to 4.0~m/s. Figure~\ref{fig:trot_bound}B shows that the fraction of time spent in bounding generally increased with both command speed and difficulty, but the rate of this increase differed across terrains. High-step, hurdle, and gap settings exhibited elevated bounding fractions even at relatively low speeds, whereas discrete and rough terrains maintained a higher proportion of trotting over comparable ranges, with bounding emerging mainly at higher velocities. These results indicate that the policy's preferred gait distribution varied jointly with velocity, terrain type, and difficulty.

The policy's performance was compared with two baselines, each restricted to a single gait, either trot or bound. As shown in Fig.~\ref{fig:trot_bound}C-i, we compared success rate, commanded-velocity tracking performance, and 1/COT across terrains and command-speed ranges: low (0-3~m/s), high (3-6~m/s), and mixed (0-6~m/s).

Figure~\ref{fig:trot_bound}C-ii summarizes the comparison between the Auto, Trot, and Bound controllers across all evaluation cases using three aggregate metrics. The best-performance rate measures how often each controller achieved the highest score, with higher values indicating better performance and the three rates summing to 100\%. The average relative performance regret is computed as the relative difference from the best score in each case, where lower values indicate that the controller stayed closer to the best performance. The worst-case normalized score is defined as the minimum normalized score across all cases, with values in the range $[0,1]$, and higher values indicating better performance. Across all three metrics, the Auto controller showed higher overall performance than both fixed-gait controllers. Definitions of the evaluation metrics and terrain difficulty levels, as well as detailed experimental settings for this section, are provided in the Supplementary Materials.

\FigureSeven

\subsection*{Training-strategy comparisons and gait-transition robustness}

This section presents simulation ablation results on complex 3D terrains, focusing on motion-prior utilization, adaptation to unseen terrains, sample efficiency, and robustness during gait transitions. We first compared our method with Adversarial Motion Priors (AMP) \cite{peng2021amp} and vanilla RL to evaluate single-gait performance on fixed terrain. We then compared our method against Hierarchical Reinforcement Learning (HRL) with a residual policy on unseen terrain to assess adaptation and transition performance under random gait switching. Figure~\ref{fig:multi_skill_evaluation}A shows the diverse terrains used for training and evaluation, including stairs, stepping stones, rough terrain, hurdles, discrete obstacles, high steps, and gaps.

Figure~\ref{fig:multi_skill_evaluation}B presents bar plots comparing our methods against AMP and a vanilla RL baseline using success rate as a robustness metric and 1/COT as a motion-efficiency metric. We compared our methods against AMP, which imitates reference behaviors using a generative adversarial imitation loss, and both our methods and AMP were trained on the same motion dataset. We also included a vanilla RL policy trained from scratch without any torque or motion priors as an additional baseline. In the main comparison, both our methods and the AMP baseline were trained with a reduced reward that included only task-related and regularization terms, whereas the vanilla RL baseline used our full reward, including all style-related components. This reduced-reward setting was constructed from our own reward formulation by retaining only the task and regularization terms and excluding the style-related rewards, following standard AMP formulations (47, 48, 53), in which the adversarial motion reward itself serves as the learned style prior. When hand-crafted style rewards were added together with the AMP adversarial reward, we observed performance degradation in cases where competing style constraints likely arose (see Supplementary Materials). The vanilla RL baseline, which lacked any motion prior, instead received our full reward to provide a strong baseline for methods without motion priors. By training separate policies for each gait-terrain pair, namely trot or bound on flat terrain, hurdles, or stepping stones, we isolated the effect of motion priors from terrain- or gait-switching capabilities. Evaluation was performed on the same terrain as training, with randomized difficulty.

Although AMP achieved a similar success rate compared to our methods on flat terrain, its COT was higher across all three terrains. We hypothesize that this degradation is related to AMP's regularization toward imitating the prior dataset, which may reduce flexibility when the terrains or motion patterns differ from those represented in the reference motions. The vanilla RL baseline achieved a comparable level of COT to our method but exhibited substantial variation in success rates across terrains when trained with a single reward setting. This indicates that terrain-specific reward tuning may be necessary to produce motion well-suited to a specific environment for vanilla RL. Our method consistently achieved lower COT and comparable success rates across all terrains, despite using motion data from flat ground only.

Next, we compared the reusability of pretrained modules from flat terrain to previously unseen 3D terrains depicted in Fig.~\ref{fig:multi_skill_evaluation}A. As illustrated in Fig.~\ref{fig:multi_skill_evaluation}C, we compared our method against an HRL framework, a widely used architecture for multi-gait locomotion \cite{hoeller2024anymal}, augmented with a residual policy \cite{silver2018residual, johannink2019residual}. For a fair comparison, we augmented the HRL baseline with a residual policy, initially trained on flat terrain and progressively adapted to complex environments via curriculum learning. During this phase, the expert policies remained fixed, whereas the residual policy was trained to handle complex terrains. As demonstrated in the learning curves shown in Fig.~\ref{fig:multi_skill_evaluation}C, our method achieved a higher velocity-tracking reward and reached higher terrain difficulty levels compared to the HRL with a residual policy baseline. The velocity-tracking reward served as the main task reward for the policy, whereas the terrain difficulty level was progressively increased through curriculum learning. Initially, the HRL with the residual approach benefited from pretrained expert policies on flat terrain. However, it struggled to adapt to unseen complex terrains and showed limited improvement, even with residual policy fine-tuning in the final stage.

During experiments, our method explored and exploited the latent space throughout training. This process was driven by an exploration bonus that progressively decayed to zero, inducing a transition from exploration to exploitation. These phases are indicated in Fig.~\ref{fig:multi_skill_evaluation}C. In addition, our method used fewer overall training samples than the HRL with residual policy baseline (Fig.~\ref{fig:multi_skill_evaluation}D), yet achieved better tracking performance because it did not require an expert pretraining phase.

Finally, we evaluate the robustness of our method under dynamic gait transitions, as shown in Fig.~\ref{fig:multi_skill_evaluation}E. In real-world settings, rapidly changing terrain features like gaps, steps, and rough surfaces demand frequent, reliable gait transitions for stable locomotion. To simulate this, we evaluate the policy from previous experiments (Fig.~\ref{fig:multi_skill_evaluation}C) under externally triggered gait switches. As illustrated in Fig.~\ref{fig:multi_skill_evaluation}E-i, dynamic transitions are simulated by randomly switching between trotting and bounding at 2~Hz with a 50\% probability, for high-frequency transitions with the same probability. Evaluation was conducted on the same terrains used during training, with difficulty levels ranging from 4 to 10.

As shown in Fig.~\ref{fig:multi_skill_evaluation}E-ii, our method achieved higher transition success rates across a range of command velocities. Transition success rate is defined as the proportion of gait transitions during which the robot remains alive throughout the transition period. As command velocity and terrain difficulty increased, success rates declined, particularly for the HRL with residual policy baseline. During experiments, the HRL baseline often failed to handle abrupt switching between separately trained expert policies, likely due to distribution mismatches. Figure~\ref{fig:multi_skill_evaluation}E-iii shows one example of the base pitch angle and gait pattern during a randomly triggered transition on rough terrain. Both methods switched from bounding to trotting by adjusting swing and stance sequences. Our method smoothly adapted the timing of these phases and gradually stabilized into a consistent trot. In contrast, the HRL with residual policy baseline failed to stabilize, resulting in stumbling and reduced survival time.

Comprehensive experimental details for this section are provided in the Supplementary Materials, including the reward formulation and training schedule, terrain generation, and difficulty parameterization.

\subsection*{Effect of perception modalities}

In high-speed agile locomotion, the robot must perceive upcoming terrain several seconds ahead to plan effectively. To achieve this, we combine a dense depth camera with a long-range 2D LiDAR. The depth camera provides dense local geometry, whereas the 2D LiDAR offers more than three times the vertical field of view, a faster update rate (40~Hz), and more reliable long-range observations. To evaluate the contribution of each sensing modality, we compared success rates across terrains at command velocities ranging from 1 to 7~m/s using only LiDAR, only the depth sensor, or both, as shown in Fig.~\ref{Fig:sensor_ablation}. The proposed method consistently achieved the highest success rates across diverse terrains. The LiDAR-only policy maintained relatively strong performance on terrains such as low stairs, rough terrain, gaps, and high steps, whereas the depth-only variant was particularly effective on hurdles and stepping-stone terrain, where precise local geometry is critical. Nevertheless, both single-modality policies underperformed compared with the proposed method. Details of the perception module and experimental settings are provided in the Supplementary Materials.

\FigureEight

\section{Discussion}
\label{discussion}

Our work extends the capabilities of quadrupedal robots to perform fast, perceptive, and adaptive multi-skill locomotion in real-world outdoor environments. We propose the APT-RL architecture that leverages a learned state representation and torque action prior derived from 2D trajectory optimization data. This prior dataset captures compact and physically grounded motion features relevant to diverse 3D locomotion behaviors. To fully leverage both state and action information from the TO dataset, we first perform representation learning followed by supervised training of an action (torque) decoder. Through this process, information from multiple gaits is embedded into a unified latent space, allowing the model to capture common structure across diverse locomotion behaviors. This pretraining process serves as the initial stage of our Action Pretrained Transformer (APT) architecture, which is later repurposed through reinforcement learning. We jointly train an auxiliary action alongside a fixed pretrained action decoder structured by a latent representation. This enables the policy to adapt to previously unseen 3D obstacle courses at high speeds, express multiple gaits in shared latent space, and generalize across diverse terrains with improved sample efficiency. Our system autonomously selects multiple locomotion gaits and corresponding motor skills by utilizing perception, terrain conditions, and command speed, supporting reliable locomotion across diverse terrain conditions. To support deployment under real-world conditions, we distill a student exteroception encoder network that utilizes ego-centric depth images and sparse long-range LiDAR into a unified exteroceptive embedding, enabling reliable terrain-aware control beyond the limited range of RGB-D cameras. We evaluated the quadrupedal robot in indoor and outdoor settings, including high steps, forest trails, leaf-covered paths, and stepped structures on a university campus.

In our real-world experiments, we encountered practical issues beyond algorithmic considerations, particularly in integrating the sensing hardware. The LiDAR produced unreliable measurements during high-speed locomotion due to large impact accelerations, often exceeding 10~g as measured by the onboard IMU, and at times this even led to sensor malfunctions or shutdown. Adding a custom 3D-printed vibration absorber between the head and the LiDAR (fig.~\ref{S_fig:head_mechanism}) reduced these impact loads, thereby keeping the exteroceptive measurements stable.

Although our current framework demonstrates dynamic and adaptive quadrupedal locomotion across diverse real-world terrains, several promising directions remain for extending its capabilities. First, although our current focus is on fore-aft locomotion primarily occurring in the sagittal plane, extending to agile behaviors such as rapid turning and lateral walking is essential for navigating more diverse and challenging terrains. Leveraging 3D motion datasets may enable the decoder to generate such behaviors and enrich the latent space representation. This extension should allow the policy to capture a broader range of dynamic skills and enhance its generalization to complex real-world locomotion tasks. Second, our policy currently supports only two gait types: trot and bound. Expanding this to include additional locomotion strategies such as pace, gallop, or crawl would enable greater adaptability in response to varying terrain conditions. Third, although this work focuses on quadrupedal locomotion, our APT-RL framework is robot-agnostic and can be extended to other embodiments, such as humanoids or legged manipulators, by generating robot-specific, trajectory-optimized motion and pretraining the decoder with a shared latent representation. Preliminary results of our method applied to different robots---such as ANYmal, Go1, and KAIST HOUND in bipedal mode---are shown in movie S4. Finally, to enable long-range autonomous missions in applications such as exploration or search and rescue, future work will integrate high-level navigation commands, long-horizon planning, and semantic understanding derived from visual perception.

\section{Materials and Methods}

In this section, we present the APT-RL pipeline, which consists of four stages. We first used trajectory optimization to generate diverse flat-terrain running motions, and then learned a TVAE-based unified latent space for trotting and bounding. Next, we trained a reinforcement learning policy that outputs latent actions decoded into joint torques via pretrained gait-specific decoders. Finally, we performed perceptual adaptation through distillation to enable control from depth images and 2D LiDAR.

\subsection*{Impulse scale-based trajectory optimization}

Unlike prior methods that rely solely on motion data from animal or human demonstrations \cite{peng2022ase, bohez2022imitate, wu2023learningamp}, our approach leverages both motion sequences and their corresponding torque profiles. This is enabled by our trajectory optimization process, which provides both state and torque information. Although the trajectories are derived from a 2D single rigid-body model, they effectively capture key locomotion features such as center of mass dynamics, angular momentum, and contact timing. The resulting dataset is physically valid, behaviorally diverse, and empirically well-suited for learning latent representations in a self-supervised learning, without relying on reinforcement learning.

To construct these trajectories, we build on the method proposed in \cite{park2015variable, park2017high} to generate quadruped motion in the 2D sagittal plane. The robot is modeled in this plane with generalized coordinates $q := (x, z, \theta)$, where $x$ and $z$ denote the horizontal and vertical positions of the body, respectively, and $\theta$ denotes the body pitch angle. To ensure the periodic cycle of the quadruped trajectory, the net impulse of the body needs to be conserved. This can be described using two momentum conservation equations.

\begin{equation}
    \begin{aligned}
    m(\dot{x}_T-\dot{x}_0) &= \int_0^{T}(F_\text{F}^x+F_\text{H}^x) dt \\
    m(\dot{z}_T-\dot{z}_0) &= -\int_0^{T} mg dt + \int_0^{T}(F_\text{F}^z+F_\text{H}^z) dt
    \end{aligned}
    \label{eq:periodic}
\end{equation}
where $F_\text{F,H}^x$ and $F_\text{F,H}^z$ are horizontal and vertical ground reaction forces (GRF) at the frontal and hind legs, respectively, $[0, T]$ is the time interval of the gait cycle, $m$ is mass of robot, and $g$ is gravitational acceleration. By modeling GRF as B\'ezier curves that satisfy Eq. \ref{eq:periodic}, the periodicity of $\dot{x}$ and $\dot{z}$ is guaranteed since momentum in the x and z directions is conserved during one gait cycle. Hence, we only consider the periodicity condition for \(z\), $\theta$, and $\dot{\theta}$ during optimization. Following the optimization technique described in \cite{park2015variable}, we obtained periodic trotting and bounding trajectories with speeds ranging from -2 to 7~m/s, body heights from 0.45 to 0.5~m, stance times from 0.13 to 0.2~s, and swing times from 0.15 to 0.21~s. Since we restrict the full model dynamics to a sagittal plane, the entire trajectory can be analytically represented using a B\'ezier curve formulation, allowing for analytical Jacobian and Hessian computation.

By minimizing the following cost function, we can obtain the initial condition \( \mathbf{q}_{\text{opt},0} := (z_0, \theta_0, \dot\theta_0) \) that induces a periodic cycle:
\begin{equation}
\label{periodic}
 f(\mathbf{q}_{\text{opt},0}) := (z_T - z_0)^2 + (\theta_T - \theta_0)^2 + \lambda_1(\dot{\theta}_T - \dot{\theta}_0)^2 + \lambda_2 \theta_{\max}^2,
\end{equation}
where \( \theta_{\max} \) denotes the maximum body pitch angle during the trajectory. The first three terms enforce periodicity, whereas the last term regularizes excessive body rotation.

To minimize this cost, we apply Newton's method using the Jacobian and Hessian of the cost function with respect to \( \mathbf{q}_{\text{opt},0} \). Specifically, let \( J = \nabla f \) and \( H = \nabla^2 f \) denote the gradient and Hessian. The update is given by:
\[
\mathbf{q}_{\text{opt,0}}^{\text{new}} := \mathbf{q}_{\text{opt},0} - \alpha H^{-1} J^\top,
\]
where \( \alpha = 1 \).

All quantities required for evaluating the cost in Equation~\ref{periodic} are extracted analytically from the B\'ezier curve. We use \( \lambda_1 = 0.5, \lambda_2 = 1.0 \) for bounding gaits, and \( \lambda_1 = 0.5, \lambda_2 = 0.0 \) for trotting gaits. The regularization term \( \lambda_2 \) is set to 1.0 only for bounding, as trotting motions typically exhibit a smaller pitch oscillation. In this case, additional penalization on the maximum pitch angle did not lead to noticeable improvements. 

The collected dataset is partitioned into two subsets: \( \mathcal{D}_\text{trot} \) and \( \mathcal{D}_\text{bound} \), each containing trajectories of state-action pairs corresponding to periodic orbits of the respective gait. In total, there are 90,000 trajectories per gait, resulting in a combined dataset of 180,000 trajectories spanning 5,560,286 time steps at 100~Hz, which corresponds to approximately 15.5~hours of motion data. The average duration of each trajectory is 0.31~seconds, with average optimization solving time \(21.582\ \mu\text{s}\). The entire process of obtaining 15.5~hours of motion data, including optimization, preprocessing, and data saving takes approximately 8 minutes.

\subsection*{Action pretrained transformer-based representation learning}

To support motor behavior generation across varying speeds and gait patterns, we perform representation learning using a TVAE. This representation integrates diverse gait types into a unified, structured latent space that can be repurposed across different tasks and environments. Based on the generated trajectories from the previous section, our representation learning process  Fig.~\ref{fig:enc_dec}B-i consists of: learning a unified state encoder for trotting and bounding by reconstructing state inputs from the motion dataset, and training torque action decoders specific to each gait using the learned latent representation.

As a first step, we train a state-only encoder that captures shared state information of trotting and bounding through reconstruction, as illustrated in Fig.~\ref{fig:enc_dec}B-i, following the approach in \cite{petrovich2021action, kang2023highly}. The encoder defines an approximate posterior distribution over motion sequences as \( q(z|s) = \mathcal{N}(z \mid \mu(s), \operatorname{diag}(\sigma^2(s))) \), where \( \mu(s), \sigma(s) \in \mathbb{R}^{16} \). A Transformer encoder predicts this distribution using tokens, where the first token corresponds to the mean \( \mu \), and the second token to the log-variance \( \log \sigma^2 \). Throughout this paper, $\log(\cdot)$ denotes the natural logarithm. The state representation includes 3 frames of proprioception information $s_t, s_{t+1}, s_{t+2}$ in $\mathbb{R}^{69 \times 3}$, including body linear and angular velocity, body height, foot height, contact information, gravitational vector of robot, joint position, joint velocity with history.

The state encoder training loss is defined as follows: $ \mathcal{L}_\text{TVAE ENC} := \mathcal{L}_\text{recon} + \lambda_\text{KL} \mathcal{L}_\text{KL} $ where $\mathcal{L}_\text{recon}:= \sum_{t=1}^3 || s_t - \hat{s}_t ||_2^2$ is defined as state reconstructing loss, $\lambda_\text{KL}$ is weighting of KL divergence loss, and \( \mathcal{L}_{\text{KL}}:= D_{\text{KL}}(q(z|s) \,\|\, \mathcal{N}(0, I)) \) is KL divergence loss, which regularizes the posterior to remain close to the unit Gaussian prior. We use a weighting coefficient of \( \lambda_{\text{KL}} = 0.1 \).

Using the shared state encoder, we can extract a latent space vector from the proprioceptive history of 2D bounding or trotting motions. Additionally, we train separate trotting and bounding torque decoders, which take the latent space vector from the shared state encoder as input and output the corresponding torque vector to be applied at each time step. We use the trained and frozen state encoder to encode three consecutive frames of proprioceptive input, \( s_t, s_{t+1}, s_{t+2} \in \mathbb{R}^{69} \), into a latent distribution. A latent vector is then sampled from this distribution and passed to the corresponding gait-specific decoder. Specifically, the trot decoder is trained using only trotting states, whereas the bounding decoder is trained using only bounding states. From each decoder's outputs, we use only the first torque vector, \( \tau_t \in \mathbb{R}^{12 \times 1} \).

The decoder is trained using the loss \(\mathcal{L}_\text{TVAE DEC} := \mathcal{L}_\text{recon} \), where the torque reconstruction loss is defined as \(\mathcal{L}_\text{recon} := \| \tau_t - \hat{\tau}_t \|_2^2,\) with \( \tau_t \) denoting the ground-truth torque and \( \hat{\tau}_t \) the predicted torque. The detailed hyperparameters used in our pretraining pipeline are provided in the Supplementary Materials. Through this representation learning process, the model acquires a gait-specific decoder trained via a structured latent action space, enabling torque-level control across multiple gaits. 

\subsection*{Reinforcement learning with action pretrained transformer}

In this section, we describe how our method reuses a pretrained decoder during the reinforcement learning stage, along with an auxiliary action. During training, the policy simultaneously explores a latent action space and learns an auxiliary action that refines the decoded torques utilizing current observations. Unlike traditional residual policies \cite{silver2018residual, johannink2019residual}, which learn an additive correction to a fixed base policy, our approach jointly learns both the latent action and its corresponding auxiliary action. Beyond pretraining distribution in flat terrain, our policy combines exteroceptive inputs from local elevation maps and proprioceptive state features to enable terrain-aware control, as illustrated in Fig.~\ref{fig:enc_dec}B-ii. The elevation data is encoded using a multi-layer perceptron (MLP) following the approach in \cite{miki2022learning}. Additionally, we adopt the concurrent training pipeline from \cite{ji2022concurrent} to train the neural velocity estimator.

We train the policy in Isaac Gym simulator \cite{makoviychuk2021isaac} using Proximal Policy Optimization (PPO) \cite{schulman2017proximal}, with a velocity-tracking reward structure modified from \cite{rudin2022learning}. Although a simpler configuration without style-specific rewards was sufficient to achieve reasonable task performance in simulation, we include additional style terms to encourage more natural and conservative obstacle negotiation and to support stable behavior on the real robot. The policy receives commands in terms of linear velocities along the x and y axes and a yaw rate. Linear velocity commands are uniformly sampled in the range of -1 to 7~m/s for the x-direction, and -1 to 1~m/s for the y-direction. The heading direction is uniformly sampled between -1.4 and 1.4 radians, and the yaw rate is set to track the commanded heading. For commanded linear x-velocity exceeding 2~m/s, the y-velocity is fixed to zero, and the heading direction is constrained to a range of -0.05 to 0.05 radians. Terrain difficulty is gradually increased via curriculum learning, with stair, stepping stones, rough, hurdle, discrete, high step, and gap, as illustrated in Fig.~\ref{fig:multi_skill_evaluation}A. The commanded linear x-velocity is restricted to the range of -1 to 1.5~m/s on high stair terrains. Details of the training setup, commanded velocity sampling, terrain configurations, and difficulty levels are provided in the Supplementary Materials.

To utilize the pretrained decoder for reinforcement learning, the actor outputs three components, as illustrated in Fig.~\ref{fig:enc_dec}B-ii-a. These components are the latent action \( a_{\text{latent}} \in \mathbb{R}^{16} \), the auxiliary action \( a_{\text{aux}} \in \mathbb{R}^{12} \), and the gait selection logit \( \text{logit}_{\text{gait}} \in \mathbb{R} \). The latent action is used as input to the pretrained decoder, the auxiliary action refines the decoded torque, and the gait-selection logit is passed through a sigmoid function \(\sigma(z) = \frac{1}{1 + e^{-z}}\) to produce a gait selection output \( a_{\text{gait}}:= \sigma(\text{logit}_{\text{gait}}) \in [0, 1] \), which indicates the preference for selecting bounding over trotting. Thus, the total action space is \( \mathbb{R}^{29} \), enabling both coarse and fine-grained motor control.

Based on the value of \( a_{\text{gait}}\), the gait is selected at 2~Hz, with latent and auxiliary actions updated at 100~Hz. If $a_\text{gait}$ is less than 0.5, the trot decoder is selected; otherwise, the bounding decoder is chosen. The selected decoder remains active for 0.5 seconds. We provide the policy with additional information, including the current gait selection output and a boolean signal that is set to true periodically at 2~Hz and false otherwise. Decoder switching is triggered only when this signal is true. The final control input is computed as $\tau_\text{input} = \tau_\text{dec} + k_p \cdot (q_\text{default} - q_t + a_\text{scale} \cdot a_\text{aux}) - k_d \cdot (\dot{q}_{t})$ where $\tau_\text{dec}:= \text{(}{a_\text{gait} < 0.5}\text{)} \cdot \text{Decoder}_\text{trot} (a_\text{latent}) + \text{(}{a_\text{gait} \geq 0.5}\text{)} \cdot \text{Decoder}_\text{bound} (a_\text{latent})$, $k_p$ is p gain of joint angle controller which is set as 80, $k_d$ is d gain of joint angle controller which is set as 2, $q_\text{default}$ is default joint angle, $q_t, \dot{q}_t$ is current joint angle and velocity, $a_\text{scale}$ is weight of auxiliary action which is set as 0.2. In conclusion, we adopt a hybrid control scheme that combines latent torque commands (learned from trajectory-optimization data) as feedforward torque setpoints with auxiliary actions (learned through reinforcement learning) interpreted as joint-space targets for a PD controller, allowing the policy to exploit both the optimized motion prior and feedback stabilization on the real robot.

Although prior work \cite{bohez2022imitate, hasenclever2020comic} incorporates encoder regularization to align policy behavior with the pretrained latent distribution, we remove this term by setting its coefficient to zero. This design allows adaptability during the RL stage, avoiding constraints from the pretraining distribution that could hinder performance in complex or unseen scenarios.

To softly constrain the latent action distribution, we apply KL regularization with respect to the standard Gaussian prior \( \mathcal{N}(0, I) \), using the following loss:
\[
\begin{aligned}
\mathcal{L}_{\text{latent action KL}}
&:= D_{\text{KL}}\!\left(
\mathcal{N}(\mu, \sigma^2)
\,\big\|\,
\mathcal{N}(0, I)
\right) \\
&= \frac{1}{2} \sum_{i=1}^{d}
\left(
\mu_i^2 + \sigma_i^2 - \log \sigma_i^2 - 1
\right),
\end{aligned}
\]
where \( d = 16 \) is the dimension of the latent action. A coefficient of \( 2.5 \times 10^{-6}\) is applied to the latent action KL regularization term.

\subsection*{Distillation of exteroception latent using teacher-student framework}

To enable real-world deployment of our controller under dynamic and high-speed locomotion scenarios, we train a student policy that operates on onboard sensor inputs, including ego-centric depth images and LiDAR data, by distilling a teacher exteroceptive encoder. Although previous approaches regress depth images to compressed latent representations \cite{miki2022learning, agarwal2023legged, kareer2023vinl, yang2023neural, zhuang2023robot, cheng2023extreme}, our method incorporates LiDAR data to overcome the limited sensing range of RGB-D cameras. The D435 depth camera, commonly used in robotic perception, provides dense spatial data within an effective range of approximately 0.3--3.0~m. However, for high-speed locomotion in outdoor environments, we require an extended sensing range beyond 3~m to ensure timely and reliable decision-making. In our case, the robot operates at over 4~m/s with gait planning at 2~Hz, requiring more than 4~m of lookahead to support at least two decision cycles and ensure timely, safe planning. To address this limitation, we incorporate a 2D LiDAR sensor, which provides sparse but accurate long-range observations. The combination of dense short-range depth data and sparse long-range LiDAR enables reliable terrain awareness at extended distances, facilitating more responsive decision-making and reducing dependence on recurrent memory mechanisms.

As illustrated in Fig.~\ref{fig:enc_dec}B-iii, we extract a latent feature vector from the depth image using a convolutional neural network (CNN) (\( z_\text{depth} \in \mathbb{R}^{32} \)), and concatenate it with the 2D LiDAR data (\( s_\text{lidar} \in \mathbb{R}^{45} \)) and proprioceptive data. Since native LiDAR raycasting was not available in the simulator at the time of our study, we trained using a 2.5D heightmap, and during deployment, applied preprocessing to match the real sensor data into the simulated format using raycasting methods. The LiDAR input is derived from a 2.5D elevation grid, where the 0.6 to 5.0~m range is discretized into 45 bins at 0.1~m resolution. We then concatenate all sensor modalities into a single composite vector, which is processed by a gated recurrent unit (GRU) encoder to produce the final perception embedding. This embedding is subsequently fed into the policy network. For student training, we follow a distillation procedure based on dataset aggregation strategy (DAgger) \cite{ross2011reduction}, combined with truncated backpropagation through time (BPTT). All other modules, such as the teacher encoder, decoder, and policy, are frozen. The student CNN, MLP, and GRU encoders are trained to regress the teacher exteroceptive latent by minimizing the mean squared error (MSE). The full deployment operates in a zero-shot manner, inferring the perception latent vector from raw depth and LiDAR inputs without any fine-tuning. Detailed hyperparameters and architecture configurations are provided in the Supplementary Materials.

\subsection*{Statistical analysis}

Quantitative results were summarized through descriptive aggregation of repeated simulated rollouts and physical robot trials. For each experimental condition, success rate was computed as the empirical proportion of evaluation units, including agents, trials, episodes, or transition events, that met the corresponding task-specific success criterion. Continuous performance metrics, including reward, COT, velocity-tracking reward, gait-selection fraction, normalized score, sample efficiency, and sensor-ablation performance, were summarized as the arithmetic mean and standard deviation over the relevant evaluation units.

Shaded regions and error bars denote standard deviation unless otherwise specified and were used to visualize variability across repeated evaluations. These descriptive summaries were not used for inferential comparison; no statistical hypothesis tests, p-values, or confidence intervals were computed. Figure-specific sample sizes and evaluation protocols, including the number of agents, trials, episodes, random seeds, command-velocity bins, terrain difficulty settings, and transition samples, are provided in the corresponding figure captions and Supplementary Materials.

Data aggregation, descriptive-statistic computation, and visualization were performed using Python with NumPy, and Matplotlib, together with MATLAB for figure processing.


\clearpage 

\bibliography{scifile} 
%
%
%
%
%
%

\section*{Acknowledgments}
The authors would like to thank Young-Ha Shin and Gijeong Kim for their work in the design of the hardware and electronics of KAIST HOUND. The authors also thank Hyunseok Kim, Sangmin Kim, Mingyu Lee, Seonsoo Kim, and Taehong Kim for their work in helping with the locomotion experiment both indoors and outdoors. The authors also thank Nguyen Trong Hieu for his contribution to implementing the bipedal mode controller for the KAIST HOUND using APT-RL. We used ChatGPT to support revising and refining the language of this paper.


\paragraph*{Author contributions:}
J.G.K., J.P., S.H., and H.-W.P. were the main contributors to the project. J.G.K., J.P., and H.-W.P. led the experimental work. J.G.K. contributed to designing the overall training pipeline. J.P. contributed to the design of the reinforcement learning and perception distillation training pipeline. J.P. contributed to implementing the controller, including the perception and policy inference pipeline. J.G.K., J.P., S.H., and H.-W.P. refined the ideas, contributed to the experimental design, and analyzed the data. T.S. contributed to implementing a sensor guard, including a mechanical vibration absorber for the LiDAR sensor. J.G.K., J.P., S.H., and H.-W.P. contributed to the writing and refinement of the paper. J.G.K., J.H.K., S.H., and H.-W.P. conceived the main idea and supervised the work.
\paragraph*{Competing interests:}
There are no competing interests to declare.

\paragraph*{Data, code and materials availability:} The data and figure generation code for this study have been deposited in the Zenodo database: \url{https://zenodo.org/records/20645964}. All data needed to evaluate the conclusions are included in the paper and Supplementary Materials.

\paragraph*{Funding:} This research was supported in part by the Challengeable Future Defense Technology Research and Development Program through the Agency for Defense Development(ADD) funded by the Defense Acquisition Program Administration(DAPA) in 2024(No.912768601), and in part by the Technology Innovation Program(or Industrial Strategic Technology Development Program-Robot Industry Technology Development)(RS-2024-00427719, Dexterous and Agile Humanoid Robots for Industrial Applications) funded By the Ministry of  Trade Industry \& Energy(MOTIE, Korea).

\vspace{-1em}

\subsection*{Supplementary materials}
Supplementary materials and methods\\
Figs. S1 to S9\\
Tables S1 to S5\\
References \textit{(82)}\\ 
Movies S1 to S4 \\
Data file S1

\MovieOneFigure

\clearpage

\renewcommand{\thefigure}{S\arabic{figure}}
\renewcommand{\thetable}{S\arabic{table}}
\renewcommand{\theequation}{S\arabic{equation}}
\renewcommand{\thepage}{S\arabic{page}}
\providecommand{\theHfigure}{\thefigure}
\providecommand{\theHtable}{\thetable}
\providecommand{\theHequation}{\theequation}
\renewcommand{\theHfigure}{S\arabic{figure}}
\renewcommand{\theHtable}{S\arabic{table}}
\renewcommand{\theHequation}{S\arabic{equation}}
\setcounter{figure}{0}
\setcounter{table}{0}
\setcounter{equation}{0}
\setcounter{page}{1}

\section*{Supplementary Materials and Methods}

\subsection*{Detailed description of states, actions, and rewards}

We can partition the policy inputs into three categories, as detailed in Table \ref{S_Tab:observation}: proprioceptive inputs, exteroceptive inputs, and gait-related inputs.

For linear body velocity, body height, foot height, and contact probability, we use the concurrent estimation strategy using ground-truth data from the simulation. We also note that ground-truth values are provided directly to the critic network during policy training.

For exteroceptive perception, we construct a latent embedding using local terrain height samples and 2D LiDAR measurements. Specifically, the height samples cover the x-direction from -0.8 to -0.2 m and 0.2 to 1.5 m, and the y-direction from -0.5 to -0.1 m and 0.1 to 0.5 m, with a resolution of 0.1 m, resulting in 210 samples in total. The 2D LiDAR samples span distances from 0.6 to 5.0 m at a resolution of 0.1 m, yielding 45 samples. These inputs are processed by the exteroceptive encoder to produce a latent embedding ($\mathbb{R}^{32}$). Note that, to improve training stability, we provided raw exteroceptive observations directly to the critic network during training, rather than passing them through the exteroceptive encoder.

Overall, the policy receives a 203-dimensional input vector composed of proprioceptive observations ($\mathbb{R}^{117}$), exteroceptive latent embeddings ($\mathbb{R}^{32}$), and gait-related information ($\mathbb{R}^{54}$). The policy output is a 29-dimensional vector, consisting of an auxiliary action ($\mathbb{R}^{12}$), an APT latent action ($\mathbb{R}^{16}$), and a gait-selection logit ($\mathbb{R}^{1}$).

We next describe the reward functions that define the training objective and shape the desired behaviors.
As summarized in Table~\ref{S_Tab:symbols}-\ref{S_Tab:reward}, the rewards are grouped into four types:
Task, Regularization, Gait-related, and Style. The form of each reward term and the notation of the variables are provided in the corresponding tables.

\subsection*{Detailed neural network design for training}

In this section, we give a detailed description of the neural network design for our training pipeline.

\subsubsection*{Pretrained APT module: encoder training}

Our latent vector from the transformer network is \( z \in \mathbb{R}^{16} \). The transformer architecture consists of a feedforward network with hidden size 64, 8 attention heads, and 2 transformer layers. We train the model using a batch size of 500, a learning rate of 0.001, and a total of 500 training epochs. All activation functions are set to the Gaussian Error Linear Unit (GeLU). 

Although many transformer-based variational autoencoder (TVAE) or representation learning methods adopt latent spaces with higher-dimensionality than ours to capture rich input structures, we set \( \dim(z) = 16 \), which was sufficient in our case. This lower-dimensional latent representation allows us to reduce computational overhead during training and inference, which is especially beneficial in reinforcement learning, where the policy needs to efficiently interpret and utilize the latent representation in real time.

The network input is proprioceptive data from Table~\ref{S_Tab:observation}, excluding the auxiliary action, resulting in a 69-dimensional input vector. The network output is identical to the input, as we train the network using a variational autoencoder (VAE) objective. 

 The encoder defines an approximate posterior distribution over motion sequences as \( q(z|s) = \mathcal{N}(z \mid \mu(s), \operatorname{diag}(\sigma^2(s))) \), where \( \mu(s), \sigma(s) \in \mathbb{R}^{16} \). Specifically, the first token of the transformer output is interpreted as the mean \( \mu \), and the second token is interpreted as the log-variance \( \log \sigma^2 \), ensuring that the model yields a valid latent distribution for sampling via the reparameterization trick. Throughout supplementary, $\log(\cdot)$ denotes the natural logarithm.

The decoder reconstructs the input from a sampled latent variable \( z \sim q(z|s) \), and the total training loss is defined as  
\( \mathcal{L}_{\text{TVAE ENC}} := \mathcal{L}_{\text{recon}} + \lambda_{\text{KL}} \mathcal{L}_{\text{KL}} \),  
where the reconstruction loss is  
\( \mathcal{L}_{\text{recon}} := \sum_{t=1}^{3} \| s_t - \hat{s}_t \|_2^2 \),  
measuring the squared error between the original and reconstructed proprioceptive inputs over a 3-step window, and the KL divergence term is  
\( \mathcal{L}_{\text{KL}} := D_{\text{KL}}(q(z|s) \,\|\, \mathcal{N}(0, I)) \),  
which regularizes the posterior to remain close to the unit Gaussian prior. We use a weighting coefficient of \( \lambda_{\text{KL}} = 0.1 \) to balance the two terms.

\subsubsection*{Pretrained APT module: decoder training}

All network parameters are identical to those in the encoder training setup. The decoder receives a latent vector sampled from the frozen state encoder, which encodes a 3-step proprioceptive history, and outputs a 12-dimensional torque vector. The decoder is trained using a standard regression loss:
\[
\mathcal{L}_{\text{TVAE DEC}} := \| \tau_t - \hat{\tau}_t \|_2^2,
\]  
where \( \tau_t \) denotes the ground-truth torque and \( \hat{\tau}_t \) is the predicted torque.

\subsubsection*{Actor and critic neural network}

The policy network is implemented as a fully connected feedforward neural network with three layers of sizes 512, 256, and 128. The actor network takes a 203-dimensional input vector and outputs a 29-dimensional action vector. The activation function used throughout the network is the Rectified Linear Unit (ReLU).

The 29-dimensional action vector consists of a 16-dimensional latent action, a 12-dimensional auxiliary action vector, and a one-dimensional gait-selection logit. The latent action is sampled from a learned Gaussian distribution and passed through a pretrained decoder to generate a torque command. The auxiliary action refines the decoded torque. The gait-selection logit is passed through a sigmoid function to compute the gait-selection output, which is used to select between two discrete gaits.

For the critic network, we replace the 32-dimensional exteroceptive latent input with the full 255-dimensional raw exteroceptive observation to allow direct value estimation from sensory information. The critic network, sharing the same architecture as the actor (except for the final layer), outputs a scalar value representing the estimated return.

Although prior work often applies encoder regularization to align policy behavior with a pretrained encoder, we omit this term by setting its coefficient to zero. This decision is based on the fact that our encoder was trained only on simplified 2D motion data, and constraining the policy to mimic this limited representation would hinder adaptation to more complex environments.

We train the policy using Proximal Policy Optimization (PPO), with each action component modeled as a Gaussian-distributed output. Entropy regularization is applied to encourage exploration, with an entropy coefficient of 0.001 for both the latent and auxiliary actions, and zero for the gait selection logit to ensure deterministic high-level decisions. Detailed parameters are summarized in Table~\ref{S_Tab:hyperparameters_ppo}.

All log-variance values are initialized to zero, corresponding to unit variance. The only exception is the gait selection logit, whose log-variance is initialized to \( \log(2) \approx 0.6931 \), corresponding to a variance of 2.

Finally, to avoid biased output at initialization, we apply mean regularization to both the auxiliary action and the gait selection logit. Specifically, we minimize
\[
\mathcal{L}_{\text{mean}} := \mathbb{E}\left[\|\mu\|_2^2\right],
\]
where $\mu$ denotes the mean of the Gaussian distribution from which the actions are sampled. This penalizes large deviations of $\mu$ from zero, thereby inducing balanced behavior and approximately equal gait selection probabilities during training. We set the regularization coefficients to $1 \times 10^{-5}$ for auxiliary actions and $0.0005$ for the gait selection logit.

\subsubsection*{Exteroceptive encoder network}

For training the teacher network, we directly use the 255-dimensional exteroceptive input as input to a fully connected neural network with three layers of size 128, 64, and 32, and an embedding size of 32. ReLU activations are applied throughout the network, whereas a $\tanh$ activation function is used for the final layer. During concurrent training of the estimation network, we omit the output of the exteroceptive encoder.

\subsubsection*{Student exteroceptive network}

To replace the exteroceptive encoder network used in the teacher model, we employed a CNN-GRU architecture that takes depth images, 2D LiDAR scans, and proprioceptive data as inputs. Similar to the approach in \cite{zhuang2023robot, cheng2023extreme}, depth images with an original resolution of 106 $\times$ 60 were downsampled to 87 $\times$ 58 and clipped. A history of 10 steps was then stacked to form the input for the CNN. The visual embeddings obtained from the CNN were concatenated with 2D LiDAR scans, proprioceptive data, and action histories, and subsequently passed through an MLP before being fed into the GRU. The GRU output was trained to match the output of the teacher exteroceptive encoder. Details of the network parameters are provided in Table \ref{S_Tab:student_extero_param}.

\subsection*{Detailed terrain types and levels for training}
Seven terrain types were used during training: stair, stepping stones, rough, hurdle, discrete, high step, and gap.
For the stair terrain, the step width was fixed at 0.3~m, whereas the step height varied from 0.05 to 0.315~m.
In the stepping stones terrain, the number of stones remained constant, whereas the size and height of each stone varied between 0.4 to 0.48~m and 0 to 0.18~m, respectively. The non-stone ground height decreased gradually along the course, ranging from 0 to 0.36~m.
For the hurdle terrain, the hurdle thickness was randomly selected from either 0.2~m or 0.3~m, and the height ranged from 0 to 0.792~m.
In the high step terrain, the thickness was sampled uniformly from 1.5 to 3~m, and the height varied from 0 to 0.9~m.
The gap width was varied from 0.1 to 1.5~m.

For the rough and discrete terrains, we used the procedural terrain generation functions provided by IsaacGym \cite{makoviychuk2021isaac}.
In the rough terrain, the minimum and maximum heights were randomly sampled from --0.06 to --0.02~m and 0.02 to 0.06~m, respectively, and a downsampled scale of 0.2 was used.
In the discrete terrain, we set the maximum height between 0 to 0.16~m, the minimum size to 0.2~m, and the maximum size to 0.8~m, with 300 randomly placed rectangular blocks. We applied the same roughness parameters as the rough terrain globally to all terrains.

All obstacle terrains were generated using a curriculum-based difficulty schedule following \cite{rudin2022learning}, where each curriculum level corresponds to a discrete terrain difficulty level. As the curriculum level increased, the terrains presented more challenging configurations.
The curriculum was divided into 10 discrete difficulty levels (levels 1--10) by linearly scaling the obstacle parameters within the ranges specified above. To encourage initial learning of flat-ground locomotion, we added two levels of flat terrain before the curriculum sequence. This design ensured that the agent acquired basic walking skills before encountering progressively more difficult obstacles.

\subsection*{Hardware implementation details}

The experiments presented in this study were conducted using KAIST HOUND, a 45~kg in-house-developed full-size quadrupedal robot. A Hokuyo UST-30LX LiDAR sensor was utilized to capture 2D LiDAR scans, and an Intel RealSense D435 provided 2D Depth images. The policy computation was executed on a ThinkStation P350 and managed using separate threads for each module: 2 kHz for low-level control, 100 Hz for the action policy (including GRU and MLP actor inference), 25 Hz for the perception module (specifically for CNN inference), 1 kHz for the estimator, 25 Hz for the camera thread, 40 Hz for the 2D LiDAR thread, and 2 kHz for data logging. For network inference, the actor policy was executed on the CPU using \textsc{ONNXRuntime}, whereas the perception latent inference was handled on the GPU using \textsc{TensorRT}. GPU inference was chosen for the perception module as CPU inference could not maintain real-time performance requirements. Torque control was implemented using ELMO Platinum Twitter 80A/80V motor drivers operating at 20 kHz.

We observed that mechanical rotational LiDAR sensors provided unreliable measurements during high-speed bounding due to the large impacts, with acceleration often exceeding 10~g as measured by the onboard IMU. These impacts frequently led to sensor malfunctions or complete failure. To mitigate this issue, a custom 3D-printed damping mechanism was implemented, effectively reducing the impact forces exerted on the 2D LiDAR sensor (Fig.~\ref{S_fig:head_mechanism}). This modification was critical for ensuring reliable sensor performance in our experiments. The vibration absorber was designed to minimize translational vibration transmitted to the LiDAR in the xz-plane, with rotational coupling suppressed. A pair of spiral springs was mounted around the LiDAR axis to suppress rotational moments, and the geometry was designed to achieve sufficiently low translational stiffness within the 10~mm displacement constraint, with high rotational stiffness retained under strict spatial constraints. The final structure was fabricated using PLA via 3D printing and validated through finite element analysis. From these experiments, we learned that for agile perceptive locomotion, it is important to mechanically isolate the IMU and exteroceptive sensors from high-frequency, body-induced vibration and impact loads and to protect them from external shocks.

\subsection*{Detailed terrain types and environments during deployments}

We conducted extensive experiments over a six-month period, spanning multiple seasons and diverse terrain types, to evaluate the robustness and adaptability of our locomotion system. During these trials, we deployed our robot across various environments, including urban campus grounds, forested areas, and indoor obstacle courses. In indoor obstacle course evaluations (Fig.~\ref{S_fig:indoor}), the robot demonstrated the ability to leap over a 60~cm-high step while achieving an instantaneous peak speed of 4.25~m/s, demonstrating its high-speed agility and dynamic obstacle negotiation capabilities.  As illustrated in Fig.~\ref{S_fig:deployment}, the robot was tested in: (top row) outdoor urban campus environments such as paved walkways and sports fields; (middle rows) forest environments with dense vegetation, uneven surfaces, exposed roots, and large natural obstacles like fallen tree branches and logs; and (bottom row) indoor obstacle courses with precisely designed platform configurations and varying obstacle heights.

\subsection*{Experimental setup and details}

\subsubsection*{Experimental setup for the analysis of gait selection rates and task performance by gait strategies across various speeds and terrains}

We generated continuous gait-fraction plots (Fig.~\ref{fig:trot_bound}B) with respect to the commanded velocity by conducting experiments across seven obstacle terrains and three difficulty levels, with velocity commands ranging from 0.5 to 4.0 m/s in 0.35 m/s intervals. Each trial lasted for 5 seconds, during which the robot traversed one to two obstacle sections. For each velocity bin, 300 trials were conducted.

For the experiments in (Fig.~\ref{fig:trot_bound}C), each terrain was constructed using the same curriculum as in the training phase of our method. For each terrain type, data were collected from 360 agents, with measurements taken for 10 seconds starting 0.2 seconds after initialization. Command velocities for the low-speed, high-speed, and mixed-speed cases were sampled uniformly from the ranges 0 to 3~m/s, 3 to 6~m/s, and 0 to 6~m/s, respectively. A single command resampling was performed 5 seconds after the start of the measurement using the same uniform distribution. The friction coefficient between the feet and the ground was set to 0.85.

Three agents, representing automatic, trot, and bound gait selection, were grouped together. We evaluated three metrics for each gait strategy: success rate, commanded-velocity tracking performance, and inverse of cost of transport (1/COT). Tracking performance was evaluated based on distance, reflecting how closely the actual motion follows the commanded displacement over an episode, similar to distance-based measures used in prior work \cite{anderson2018evaluation}, and is given by 
\begin{equation}
    \begin{aligned}
    \frac{d_\text{des} - \left| d_\text{des} - d_\text{real} \right|}{d_\text{des}}
    \end{aligned}
    \label{tracking_avg}
\end{equation}
where $d_\text{des}$ represents the desired distance, calculated as $\left\|v_\text{cmd,xy}\right\| \cdot t_\text{survive}$, with $v_\text{cmd,xy}$ and $t_\text{survive}$ denoting the command linear velocity and the time survived in the episode, respectively. Since $v_\text{cmd,xy}$ is fixed during each episode, the desired distance increases proportionally with the time survived. The term $d_\text{real}$ corresponds to the actual distance traveled by the robot during the episode. The success rate is defined as the number of agents that remain operational at the end of the episode divided by the total number of agents. 
We define the average mechanical cost of transport as
\begin{equation}
    \begin{aligned}
        COT_\text{mech}^\text{avg} = \frac{\sum\limits_{i} E_i}{mg \cdot \sum\limits_{i} d_{\text{real},i}} \\
    \end{aligned}
    \label{cot_mech}
\end{equation}
where
\begin{equation}
    \begin{aligned}
        E_i =  \sum\limits_{t=1}^{T_i}  \left(\sum_{\text{actuators}}[ \boldsymbol{\tau}_t^{(i)}  \boldsymbol{\dot{q}}_t^{(i)}]^+\right)  \Delta t
    \end{aligned}
    \label{cot_mech2}
\end{equation}
Here, for each agent $i$, $\boldsymbol{\tau}_t^{(i)}$ and $\boldsymbol{\dot{q}}_t^{(i)}$ denote the joint torque and joint velocity at time step $t$. $\Delta t$ is the time step duration, $T_i$ is the number of time steps in the episode, $d_{\text{real},i}$ is the actual distance traveled during the episode, and $mg$ denotes the total weight of the robot. The operator \([\cdot]^+\) denotes the positive part, ensuring that only energy-consuming (non-regenerative) power contributions are considered.

\subsubsection*{Experimental setup and training details for comparison with motion-prior-based algorithms}

To compare with motion-prior-based methods on our dataset, we adopted the AMP \cite{peng2021amp} framework as a baseline. For each gait, both methods were trained using the corresponding motion dataset, either bounding or trotting, generated via trajectory optimization on 2D flat terrain. Specifically, we trained separate policies on flat, hurdle, and stepping stones terrains using the bounding and trotting datasets to track commanded velocities ranging from 1 to 7~m/s. The discriminator received the following state information: joint positions and velocities, and base linear and angular velocities. We excluded all roll-related joint states and base height information from the discriminator inputs to make it more difficult to distinguish between dataset motions and task executions.

AMP was implemented following the structure of \cite{wu2023learningamp}, and was built on top of the code provided in \cite{escontrela2022adversarial}. The AMP reward coefficient was set to 0.1, and the task-reward linear interpolation function used a coefficient of 0.3. We aligned our overall training setup with \cite{wu2023learningamp} as closely as possible. Our method and the AMP baseline were both trained using only the regularization and task rewards listed in Table~\ref{S_Tab:reward}, which were largely consistent with those used in \cite{wu2023learningamp}. For the AMP baseline, the latent-regularization term listed in Table~\ref{S_Tab:reward} was excluded. For the AMP baseline, privileged information described in \cite{wu2023learningamp} was provided as policy inputs in order to follow the original training procedure. For our proposed method, the estimator network was jointly trained with the policy using the same architecture employed for real-world deployment.

We also included a vanilla RL policy trained without action or motion priors for comparison. Its reward configuration was identical to that of our method except that all gait-related and latent-action-related reward terms were removed. During policy training, the estimator network and exteroceptive encoder were jointly optimized with the actor.

In terms of experimental setup, data were collected from 300 agents, each sampling commanded velocities uniformly from 1 to 7~m/s. Obstacles used during evaluation were randomly selected from specified difficulty intervals: hurdle heights ranged between 0.22 to 0.792~m, whereas stepping stones had sizes between 0.4 to 0.48~m and heights up to 0.18~m. Each episode lasted 10 seconds and was repeated 10 times to calculate the average performance. For evaluation, the friction coefficient between the feet and the ground was set to 0.85. For evaluation, we used the same definitions of success rate and mechanical cost of transport (COT) as described in the Supplementary subsection on gait selection rates and task performance.

\subsubsection*{Training setup for full obstacle terrain training comparison with hierarchical reinforcement learning with residual}

To ensure a fair comparison, the low-level policies in the Hierarchical Reinforcement Learning (HRL) implementation adopted the same structure as our proposed method, using a single gait decoder for each gait. Trot and bound gait policies were independently trained on flat terrain without gait-related reward terms, and the high-level controller was then trained following the same procedure as our unified policy. The proposed method also reused the pretrained decoder, which was trained on a 2D flat-terrain trajectory optimization dataset, without any additional finetuning. Both methods are trained to track a commanded forward velocity range of -1 to 7~m/s. 

\subsubsection*{Experimental setup for robustness under random gait transition}

The random gait transition robustness experiment (Fig.~\ref{fig:multi_skill_evaluation}E) was conducted by training both methods on the full obstacle terrain, using the same setup as in Fig.~\ref{fig:multi_skill_evaluation}C. Random gait switching was applied at 2~Hz with a probability of 50\% on the entire obstacle terrain (Fig.~\ref{fig:multi_skill_evaluation}A), as demonstrated in Fig.~\ref{fig:multi_skill_evaluation}E-i. 

Commanded velocities were set from 1 to 7~m/s, with early termination disabled to assess generalization under out-of-distribution (OOD) conditions. We collected 75,000 samples until each agent was terminated. For each agent, the episode lasted up to 10~seconds, allowing a maximum of 20 transitions. We excluded samples in which the agent was terminated within the initial 0.5~seconds, which corresponds to the time of the first random transition.

The grid plot corresponding to Fig.~\ref{fig:multi_skill_evaluation}E-ii is illustrated in Fig.~\ref{S_fig:random_grid_plot}. The numbers in the grid indicate the number of samples used to calculate the transition success rate with respect to each commanded velocity and terrain level.
Terrain difficulty for evaluation was randomly sampled from curriculum levels 4 to 10. The terrain level at the starting point was fixed to level 3.

For the analysis of base pitch angle trajectory and gait pattern in Fig.~\ref{fig:multi_skill_evaluation}E-iii, we collect samples on rough terrain at a commanded velocity of 4~m/s. At 1.5~seconds after the episode begins, we enforce a gait transition.

\subsubsection*{Experimental setup for Perception Modality Ablations}

In this experiment, all policies were trained using the same teacher, with the perception strategy as the only difference. We compared three student policies. The proposed method used both depth camera and LiDAR inputs, the LiDAR-only variant removed the CNN branch, and the depth-only variant excluded LiDAR inputs. The experimental setup consisted of seven environments: Low stair, Rough, Gap, Hurdle, High step, Stepping stones, and Discrete. Terrain difficulty for evaluation was randomly sampled from curriculum levels 4 to 10. The terrain level at the starting point was fixed to level 3. For each experiment, 100 agents were evaluated per trial over 10 trials.

\subsection*{Ablation study: Architectural components}

We conducted ablation studies to analyze the contribution of each component with the same training setup as our deployed policy. We compared the full method with three ablated variants that selectively disabled the auxiliary action, the shared latent structure, or RL-based latent-action optimization.
    
The full method, shown by the cyan curve in Fig. \ref{S_fig:flat_ablation}, used gait-specific torque decoders implemented through torque control, a unified latent action space shared across the decoders, RL-based policy optimization for latent-action selection, and an auxiliary action. Our full method clearly achieves both the fastest learning and highest final performance, with reward rising very steeply at the beginning and surpassing all other variants early in training.

The variant without the auxiliary action, shown by the blue curve, uses decoders with a shared unified latent action space together with RL-based policy optimization for latent action, but does not use the auxiliary action. Since the TO prior data are constructed under a 2D dynamics approximation, the abduction/adduction target poses are set to zero in the low-level PD (Proportional-Derivative) controller. In this case, we observe that the velocity-tracking reward remains lower overall throughout training compared to our full method, as shown in Fig.~\ref{S_fig:flat_ablation}. We hypothesize that, without auxiliary action, the policy lacks the ability to compensate for 3D effects that are absent in the TO dataset, which is consistent with the observed performance degradation.

The variant without a shared latent space, shown by the red curve, separately trained trotting and bounding experts that do not share a unified latent space, and then used hierarchical RL to build an expert policy capable of utilizing both gaits. Although this variant shows better performance than the other comparison variants, its performance remains lower than the full method and does not exhibit noticeable improvement during training. In addition, it still requires pretrained policies for each gait separately. Although the expert policy can employ both gait controllers, it often fails during transitions between gaits.

The brown curve shows the variant without RL-based policy optimization for latent actions. In our full method, we use RL-based policy optimization to select latent actions from observations. To further evaluate the effectiveness of these policy-generated latent actions, we compared our policy against a baseline that replaces the latent actions produced by the RL policy with latent actions obtained from a pretrained encoder trained on 2D motion sequences, which also maps observations to latent actions. This baseline showed lower performance throughout training compared to our full method, similar to the variant without auxiliary action but with larger deviations, as shown in Fig.~\ref{S_fig:flat_ablation}. This observation is consistent with RL-based policy optimization for latent action being important for effectively reusing the learned latent space.

The results show that our method, which includes TO--based learning of a unified latent action space for the decoders, RL-based latent action reuse, and auxiliary action, achieves the most stable and consistent training performance. Overall, the ablation indicates that all components contribute meaningfully to performance, and that the complete configuration is required to obtain the highest and most stable velocity-tracking and success-rate performance. Adding both the auxiliary action and the latent RL policy consistently improves performance over simpler variants, with the latent RL policy in particular reducing training deviation, and the unified latent space between trotting and bounding torque decoders provides additional gains when both gaits are employed.

\subsection*{Ablation study: Effect of dataset size for representation learning}
Throughout training, the number of samples for encoder training was selected to ensure sufficient coverage of the motion manifold across command velocities and gait patterns. In our setup, each gait decoder is trained from samples collected via trajectory optimization over command velocities from -2 m/s to 7 m/s, using 1 m/s intervals with 10,000 short trajectory segments per interval. In total, there are 90,000 trajectories per gait, resulting in a combined dataset of 180,000 trajectories spanning 5,560,286 time steps at 100~Hz, which corresponds to approximately 15.5~hours of motion data. The average duration of each trajectory is 0.31~seconds, with average optimization solving time \(21.582\ \mu\text{s}\). The entire process of obtaining 15.5~hours of motion data, including optimization, preprocessing, and data saving, takes approximately 8 minutes.

To verify whether such dense sampling is necessary, we trained the state-representation encoder used in our representation learning stage with 1\%, 10\%, 50\%, and 100\% of the dataset. As shown in Fig.~\ref{S_fig:dataset_ablation}, using only a small fraction (1\%) of the dataset leads to slower convergence and noticeably higher total loss. We also observe that as the data ratio increases, the final loss decreases consistently. The state encoder training loss is the same as the previous section: $ \mathcal{L}_\text{TVAE ENC} := \mathcal{L}_\text{recon} + \lambda_\text{KL} \mathcal{L}_\text{KL} $ with a weighting coefficient of \( \lambda_{\text{KL}} = 0.1 \).

\subsection*{Ablation study: Comparison with AMP under Our Full Reward Formulation}

As an additional control experiment for supplementary materials, we trained AMP with the full reward formulation. In this setting, AMP received the same task, regularization, and style-related rewards as our method, except for terms that are specific to our framework and not applicable to AMP, such as latent-action and gait-selection terms. The adversarial AMP reward was then added as the data-driven motion-prior term. This differs from the main text experiments, where we used a shared reward setting, which was composed of task-related objectives and regularization terms, with style-related terms omitted.

This full-reward AMP variant did not yield stable or reliable locomotion, particularly when trained with the trotting reference dataset. In repeated trials, the learned policy showed poor command tracking, excessive dependence on reference-like motion patterns, or loss of coherent gait structure. The result in Fig.~\ref{S_fig:amp_full_comparison} was obtained using the same reward coefficients, training setup, and evaluation protocol as the other baseline experiments. The degradation was most evident on hurdle terrain, where the trotting-data AMP policy produced a near-zero success rate.

These observations indicate that combining the AMP discriminator with additional hand-crafted style rewards can overly constrain the policy when the training environment requires motions outside the reference dataset. The discriminator favors behaviors close to the flat-ground motion distribution, whereas style-related terms further shape swing, contact, and foot-clearance patterns. For obstacle traversal, however, the policy must adapt these motion patterns beyond the range covered by the flat-terrain data. This effect is more pronounced for the trot dataset, whose kinematic envelope is less aligned with the motions required for obstacle traversal, whereas the bound dataset, whose synchronized paired-leg motion provides larger body clearance and flight phases, is more compatible with obstacle-crossing behaviors.

\subsection*{Ablation study: Feedforward torque contribution}

\subsubsection*{Scaling the feedforward torque}
To quantify the contribution of the pretrained torque prior $\tau_\text{dec}$, we 
scale its contribution from 0\% to 100\% at deployment in simulation (3 seeds, 300 
environments, command velocity $[0, 6]$ m/s, high-step terrain levels 2--5). As 
shown in Fig.~\ref{S_fig:ablation_ff_torque}, removing $\tau_\text{dec}$ entirely 
causes the success rate to collapse to 2.9\%, whereas restoring it to full scale 
recovers performance to 94.6\%. Velocity tracking reward degrades from 1.384 to 
0.751 as $\tau_\text{dec}$ is removed. We note that the apparent increase in 
1/COT at low scales is misleading: robots simply fail to track high-speed commands, 
artificially lowering energy consumption. These results confirm that the pretrained 
torque prior $\tau_\text{dec}$ is the dominant and non-redundant component of the 
control signal, carrying the gait-cyclic dynamics learned from ground reaction 
force (GRF)-based trajectory optimization (TO) data, and cannot be replaced by 
the auxiliary PD term alone.

\subsubsection*{Equivalence with PD-target formulation}
Our hybrid controller can be algebraically re-expressed as a PD controller with a 
shifted reference. Specifically, setting $\tau_\text{input} = k_p \cdot (q_\text{default} - q_t + q_\text{ref}) - k_d \cdot \dot{q}_t$ equal to our original control law and solving for $q_\text{ref}$ yields $q_\text{ref} = a_\text{scale} \cdot a_\text{aux} + \tau_\text{dec} / k_p$, which aligns with the conventional PD-only form used in the legged-locomotion literature. The two forms are not exactly equivalent in practice due to differing 
control rates (100~Hz for $\tau_\text{dec}$ and $a_\text{aux}$ vs.\ 400~Hz for 
the PD loop), but as shown in Fig.~\ref{S_fig:ablation_pd_target}, they yield 
virtually identical performance using the same trained policy (success rate 
94.3\% vs.\ 94.3\%, velocity tracking reward 1.383 vs.\ 1.382, 1/CoT 2.394 
vs.\ 2.384). The choice between them is therefore primarily a matter of 
representation, not capability.

Taken together, these results empirically justify the feedforward-torque formulation in two ways. The pretrained torque prior $\tau_\text{dec}$ 
is essential and carries the dominant locomotion signal, as performance collapses 
without it. Converting $\tau_\text{dec}$ into an equivalent PD target 
yields virtually identical performance, confirming that the two forms are 
interchangeable in practice.

\newpage

\begin{figure*}
\centering
	\includegraphics[width=1.0\linewidth]{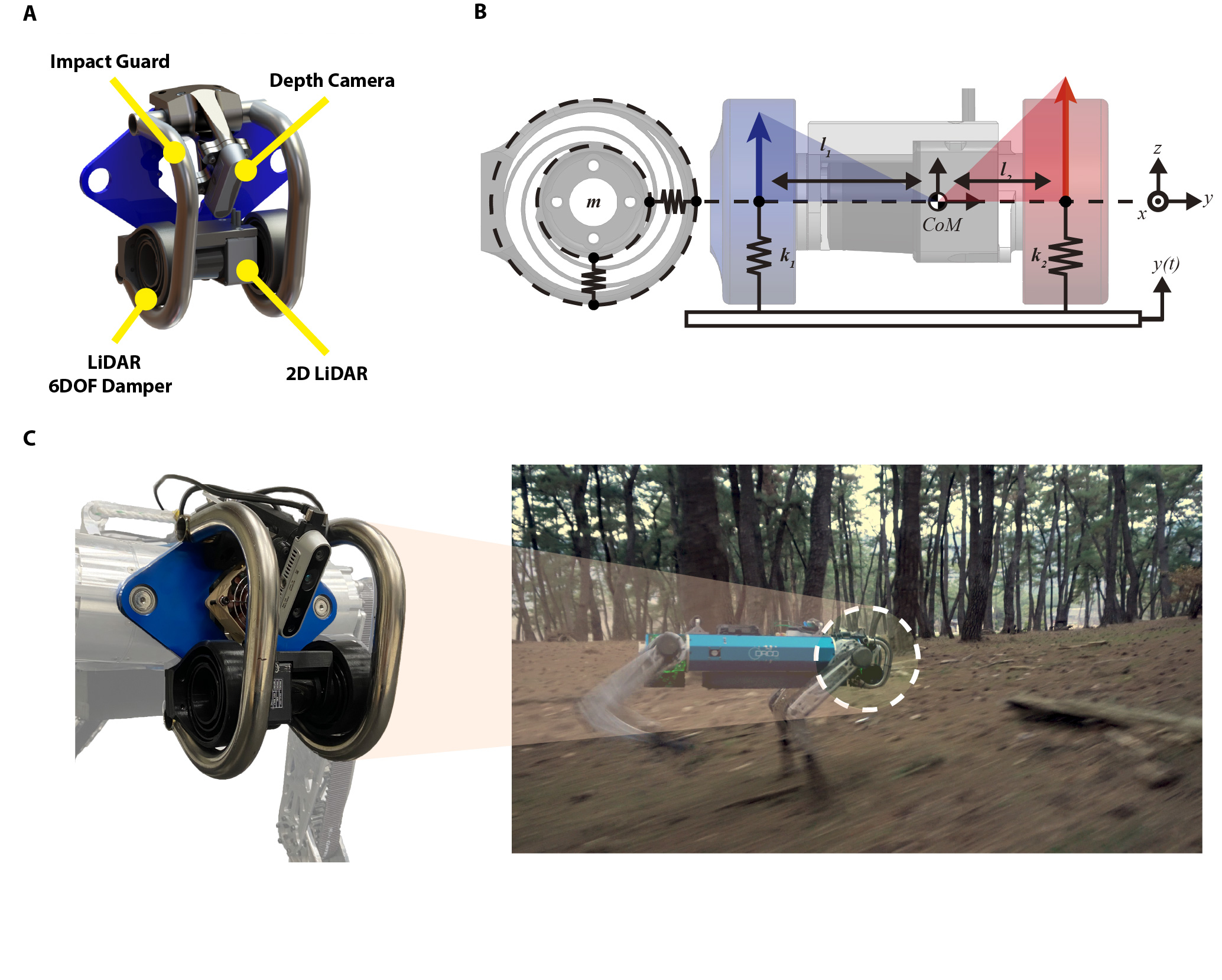}
    \caption{\textbf{Perception hardware module.} (\textbf{A}) Perception hardware module equipped with a depth camera, 2D LiDAR, custom-designed LiDAR vibration absorber, and impact guard. (\textbf{B}) Approximate mass-spring model of the vibration absorber. (\textbf{C}) Perception module mounted on the robot with all components assembled and deployed in hardware. 
    } 
    \label{S_fig:head_mechanism}
\end{figure*}

\begin{figure*}[t]
	\centering
	\includegraphics[width=\linewidth]{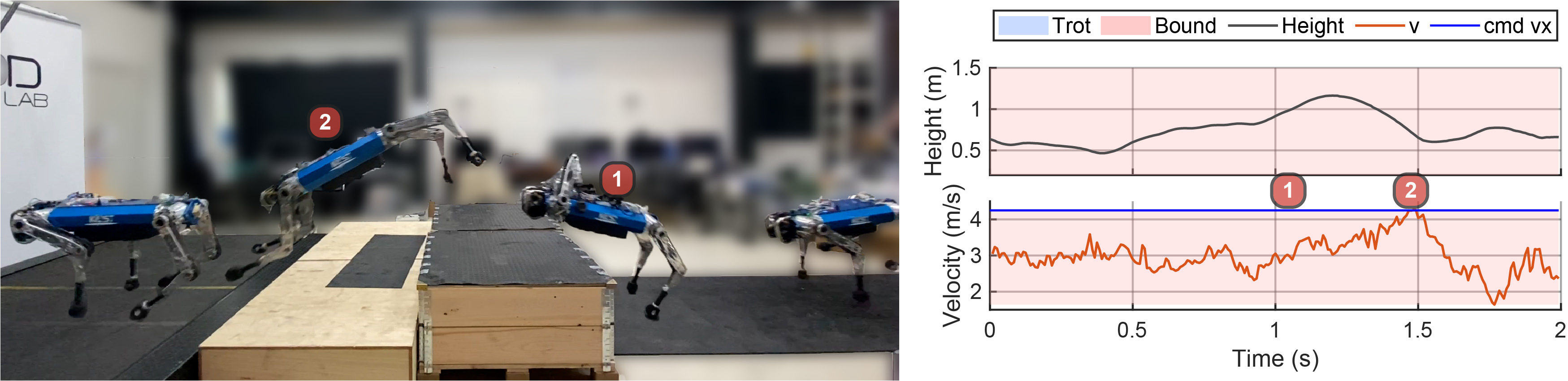}
    \caption{\textbf{Camera tracked velocity during traversal of the 60~cm obstacle.}} 
    \label{S_fig:indoor}
\end{figure*}

\begin{figure*}
\centering
	\includegraphics[width=\linewidth]{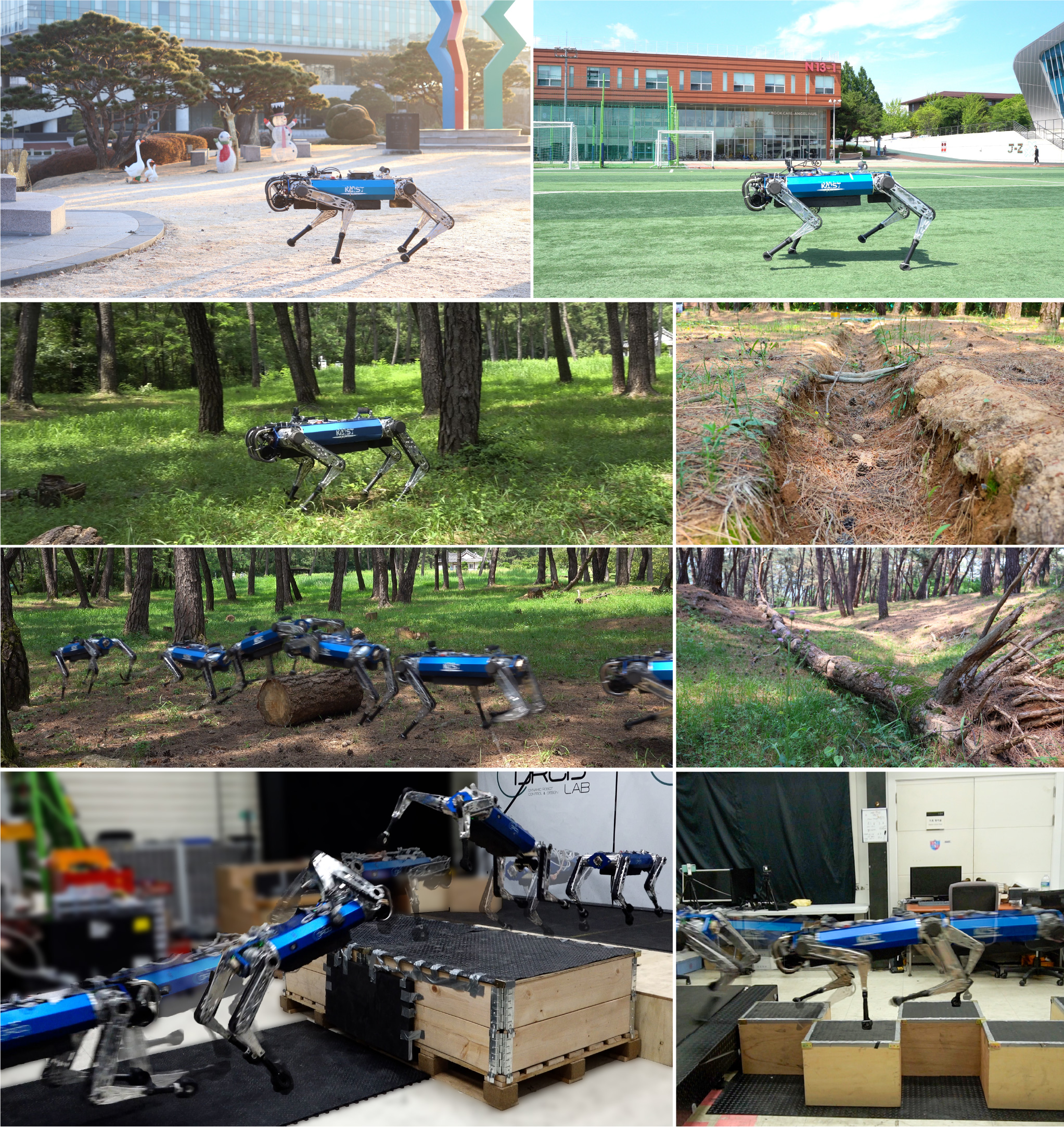}
    \caption{\textbf{Various terrains during experiments.}}
    \label{S_fig:deployment}
\end{figure*}

\newpage

\begin{figure*}
\centering
	\includegraphics[width=0.6\linewidth]{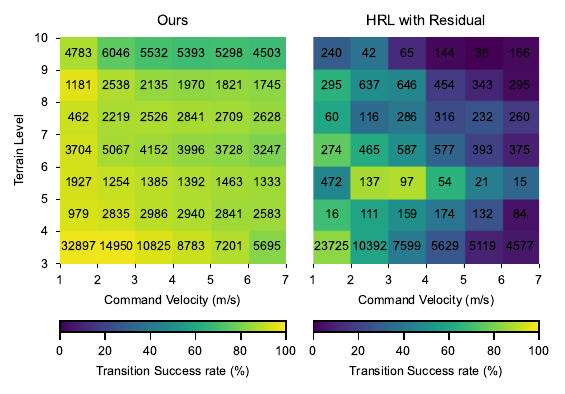}
    \caption{\textbf{Transition success rate grid plot for robustness under random gait transition.}
    } 
    \label{S_fig:random_grid_plot}
\end{figure*}

\begin{figure*}
\centering
	\includegraphics[width=0.6\linewidth]{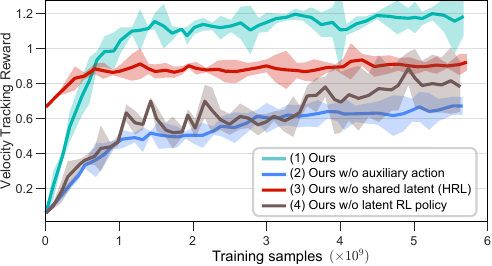}
    \caption{\textbf{Ablation study of the architectural components.} Velocity tracking on flat terrain with HRL variants and ablations of auxiliary actions and latent input generation. Shaded regions indicate standard deviation across three random seeds.
    } 
    \label{S_fig:flat_ablation}
\end{figure*}

\begin{figure*}
\centering
	\includegraphics[width=0.5\linewidth]{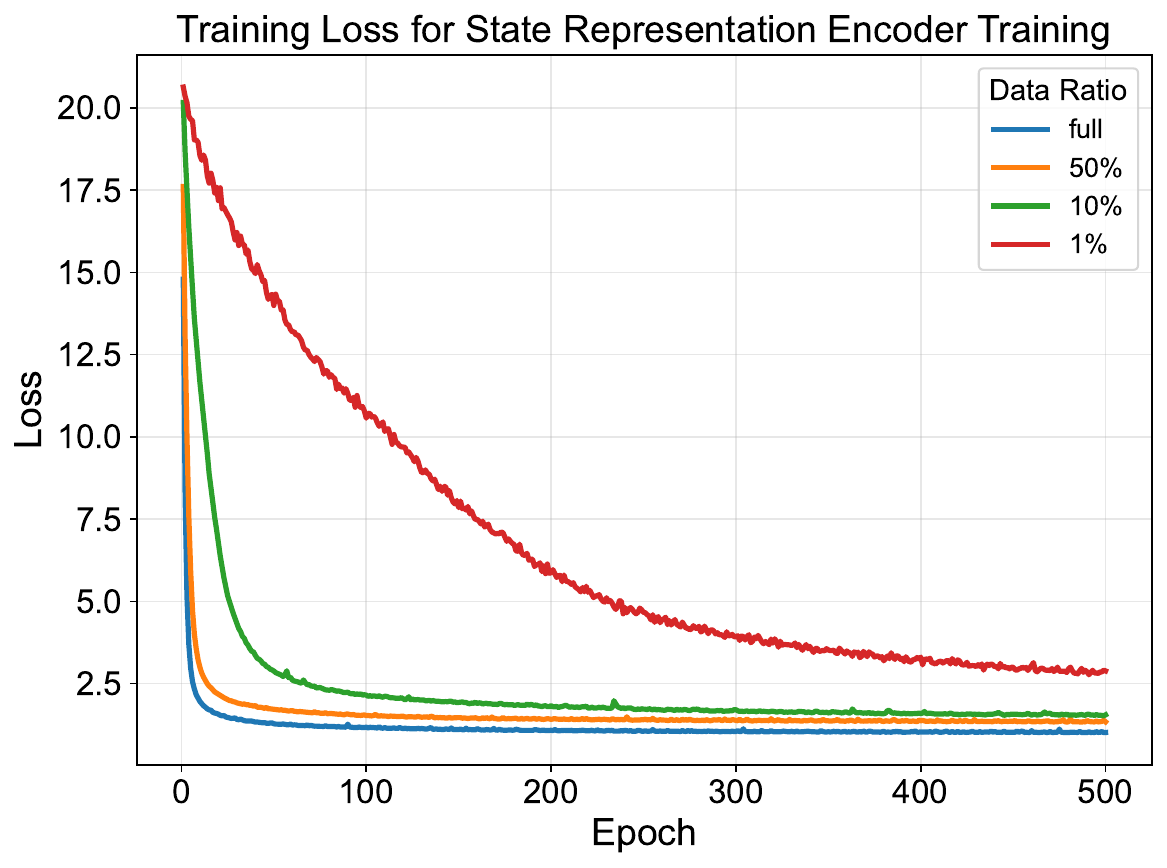}
    \caption{\textbf{Dataset Size Ablation Study.
    } Training loss of the state-representation encoder with 1\%, 10\%, 50\%, and 100\% of the dataset, showing that larger datasets yield consistently lower final loss.
    }
    \label{S_fig:dataset_ablation}
\end{figure*}

\begin{figure*}
\centering
	\includegraphics[width=1.0\linewidth]{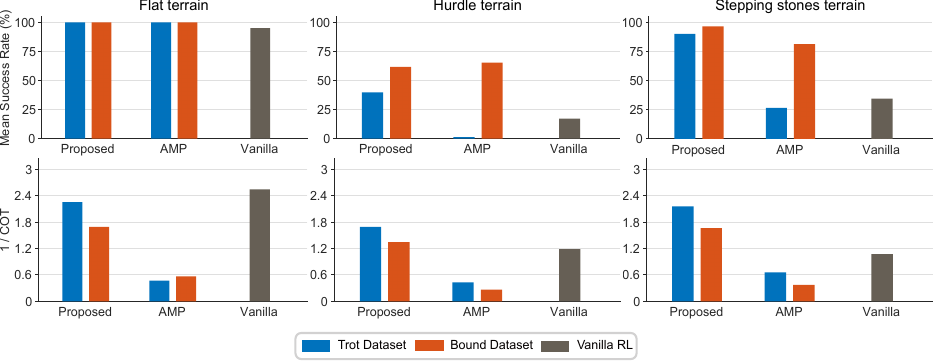}
    \caption{\textbf{Comparison of algorithms differing in prior utilization under the proposed reward configuration including style-related terms.} AMP and vanilla RL were trained with the proposed task, regularization, and style-related reward terms, excluding method-specific terms not defined for each baseline. The AMP adversarial reward was additionally included for AMP.}
    \label{S_fig:amp_full_comparison}
\end{figure*}

\begin{figure*}
\centering
    \includegraphics[width=\linewidth]{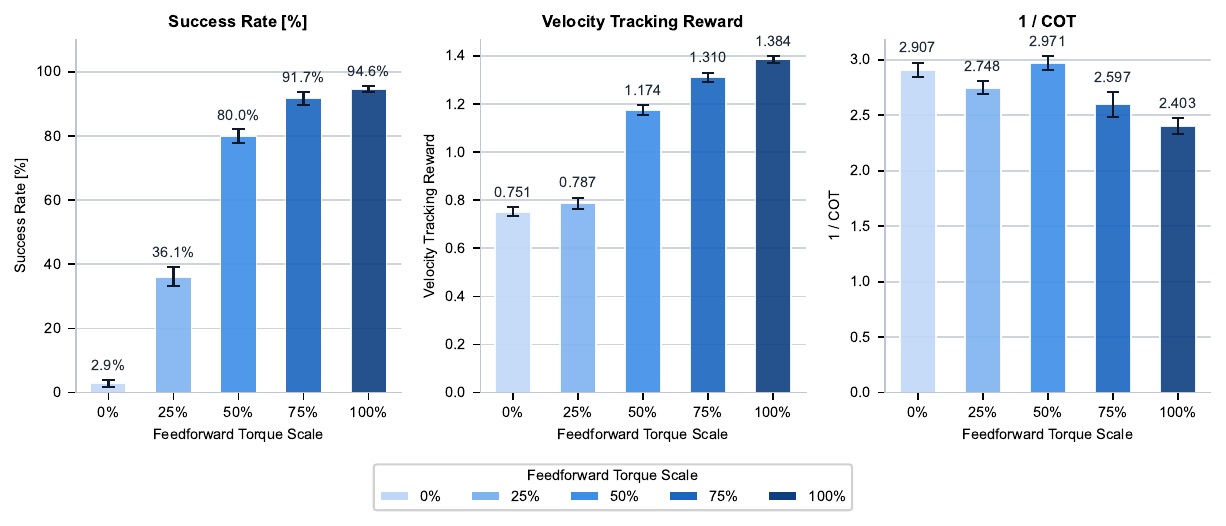}
    \caption{
    \textbf{Feedforward torque ablation.} 
Results are reported as mean $\pm$ std over 3 random seeds, with 300 environments per seed; commanded velocities were sampled from $[0, 6]$ m/s, and high-step terrain levels 2--5 were used. Scaling $\tau_\text{dec}$ 
from 0\% to 100\% monotonically improves success rate (2.9\%$\to$94.6\%) and velocity 
tracking reward (0.751$\to$1.384). The apparent 1/CoT drop at full scale reflects 
higher locomotion speed rather than reduced efficiency.
    }
    \label{S_fig:ablation_ff_torque}
\end{figure*}

\begin{figure*}
\centering
    \includegraphics[width=\linewidth]{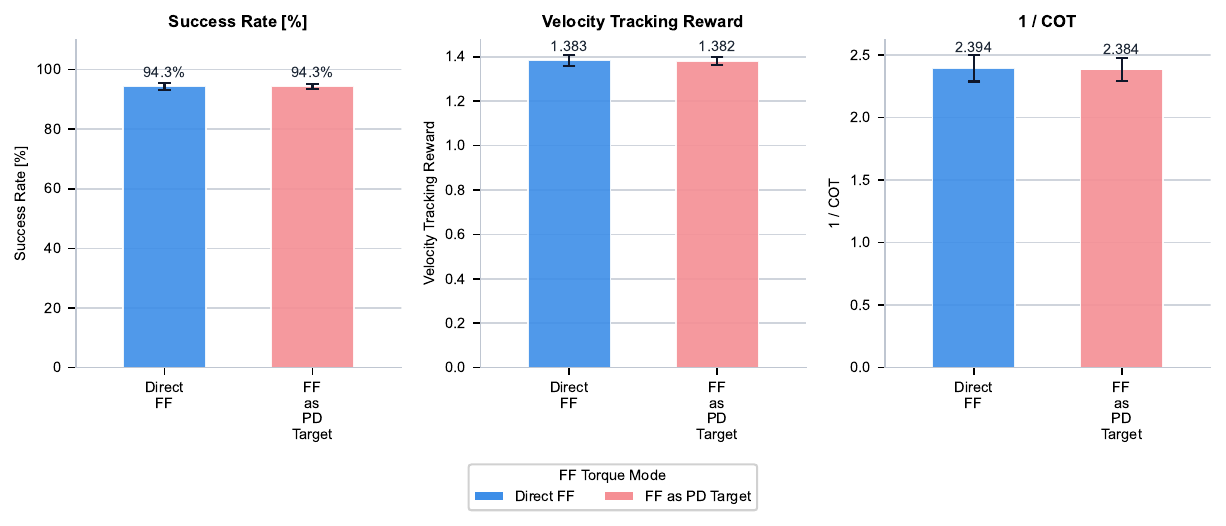}
    \caption{\textbf{Comparison between direct feedforward torque control and the equivalent PD-target formulation.} 
Results are reported as mean $\pm$ std over 3 random seeds, with 300 environments per seed; commanded velocities were sampled from $[0, 6]$ m/s, and high-step terrain levels 2--5 were used. Both forms yield virtually identical success rate (94.3\%), velocity tracking reward (1.383 vs.\ 1.382), and 1/CoT (2.394 vs.\ 2.384), confirming that the choice between them is a matter of representation, not capability.}
    \label{S_fig:ablation_pd_target}
\end{figure*}

\clearpage

\begin{table*}[t]
\centering
\caption{Observation types and dimensions.}
\vspace{1.0em}
\label{S_Tab:observation}
\renewcommand{\arraystretch}{1.20}
\begin{tabular}{llc}
\hline
\textbf{Observation Type} & \textbf{Input} & \textbf{Dim.} \\
\hline
\textbf{Proprioception} & linear body velocity estimation & 3 \\
               & angular body velocity & 3 \\
               & body height estimation & 1 \\
               & foot height estimation & 4 \\
               & contact probability estimation & 4 \\
               & command velocity & 3 \\
               & projected gravity vector & 3 \\
               & joint position & 12 \\
               & joint velocity & 12 \\
               & joint position history (2 time steps ago) & 12 \\
               & joint velocity history (2 time steps ago) & 12 \\
               & joint torque & 12 \\
               & joint torque history (2 time steps ago) & 12 \\
               & auxiliary action & 12 \\
               & auxiliary action history (2 time steps ago) & 12 \\
\hline
\textbf{Exteroception} & local height samples & 210 \\
              & 2D LiDAR height samples & 45 \\
\hline
\textbf{Gait related} & selected gait history & 10 \\
              & gait selection output history & 10 \\
              & latent action for APT decoder & 16 \\
              & latent action history for APT decoder (2 time steps ago) & 16 \\
              & is current gait selected & 1 \\
              & current not selected gait output & 1 \\
\hline
\end{tabular}
\end{table*}

\begin{table*}[t]
\centering
\caption{List of Symbols and Notations for Rewards}
\vspace{1.0em}
\label{S_Tab:symbols}
\renewcommand{\arraystretch}{1.20}
\begin{tabular}{ll}
\hline
\textbf{Symbol} & \textbf{Description}\\
\hline
$v^{\text{cmd}}_{b,xy},\ \bar v_{b,xy}$ & Commanded and current average horizontal body velocity \\
$w^{\text{cmd}}_{b,z},\ \bar w_{b,z}$ & Commanded and current average yaw body velocity \\
$a_{\text{latent}}$ & Latent action (input to the pretrained gait decoder) \\
$a_{\text{aux}}$ & Auxiliary action used to refine the decoded torque \\
$a_{\text{gait}}$ & Gait-selection output: $a_{\text{gait}} < 0.5$ (trot), $a_{\text{gait}} \geq 0.5$ (bound) \\
$q,\ \dot{q},\ \ddot{q}$ & Joint positions, velocities, and accelerations \\
$q_{\text{default}}$ & Nominal joint positions \\
$R_{\text{phase}},\ K_t$ & Motor phase resistance (0.57) and torque constant (0.72) \\
$\tau$ & Joint torque \\
$T_{\psi}^s,\ T_{\psi}^l$ & Gait selection buffer size (5 steps, 2.5 s / 10 steps, 5 s) \\
$\psi,\ \bar\psi_{1:t}$ & Current gait selection ($\psi=0$ trot, $\psi=1$ bound) and its mean over time steps $1$ to $t$ \\
$c_f,\ h_f,\ v_f,\ F_f$ & Contact indicator, height, linear velocity, and contact force for foot $f$ \\
$T_{a,i},\ T_{s,i}$ & Swing and stance time for the $i$th leg \\
\hline
\end{tabular}
\end{table*}

\begin{table*}[t]
\caption{Reward functions}
\vspace{1.0em}
\label{S_Tab:reward}
\centering
\scriptsize
\setlength{\tabcolsep}{6pt}
\renewcommand{\arraystretch}{1.20}
\begin{tabular}{p{1.7cm} p{3.0cm} p{8.2cm} c}
\hline
\textbf{Type} & \textbf{Name}  & \textbf{Equation} & \textbf{Weight} \\
\hline
\textbf{Task} 
& Linear velocity tracking  
& \( r_\text{lv} =
\begin{cases}
\exp\!\left(-\left\|\dfrac{v^{\text{cmd}}_{b, xy}-\bar v_{b, xy}}{v^{\text{cmd}}_{b,x}}\right\|_2^2/0.25\right), & v^{\text{cmd}}_{b, x}>0.1,\\[0.2em]
\exp\!\left(-\|v^{\text{cmd}}_{b, xy}-\bar v_{b, xy}\|_2^2/0.25\right), & \text{otherwise}
\end{cases}
\)
& 1.5\\

& Angular velocity tracking  
& \( r_\text{av} = \exp\!\left(-\|w^{\text{cmd}}_{b, z}-\bar w_{b,z}\|_2^2/0.25\right) \)
& 0.5\\
\hline

\textbf{Regular-\newline ization}
& Latent action smoothness
& \( r_\text{la}=\|a_{\text{latent},t}-a_{\text{latent},t-1}\|_2^2 
   + 0.15\|a_{\text{latent},t}-2a_{\text{latent},t-1}+a_{\text{latent},t-2}\|_2^2 \)
& -0.01 \\

& Auxiliary action smoothness
& \( r_\text{jts}=\|a_{\text{aux},t}-a_{\text{aux},t-1}\|_2^2
  +\|a_{\text{aux},t}-2a_{\text{aux},t-1}+a_{\text{aux},t-2}\|_2^2 \)
& -0.01 \\

& Auxiliary action scale
& \( r_\text{as}=\sum_{i=1}^{12} a_{\text{aux},i}^2 \)
& -0.007 \\

& Joint torque
& \( r_\tau = (R_\text{phase}/K_t^2)\sum_{i=1}^{12}\tau_i^2 \)
& \(-1.0\cdot10^{-7}\) \\

& Joint acceleration
& \( r_\text{acc}=\sum_{i=1}^{12} \ddot q_i^2 \)
& \(-1.0\cdot10^{-5}\) \\

& Termination penalty
& \( r_\text{ter}=\mathbf{1}[\text{base coll.}\lor \text{leg coll.}\lor \text{Base Roll/Pitch/Joint limit}] \)
& -1.5 \\
\hline

\textbf{Gait related}
& Gait diversity
& \( r_\text{gd} = c_\text{gd} (\bar\psi_{1:T_{\psi}^l}-0.5)^2,\quad c_\text{gd}=1-\tfrac{\text{epoch}}{50000} \)
& -15 \\

& Gait change penalty
& \( r_\text{gc}=c_\text{gc}\sum_{t=0}^{T_{\psi}^s}(\psi_{t+1}-\psi_t)^2,
\;
c_\text{gc}=\big(1-\tfrac{\text{epoch}}{50000}\big)+\tfrac{1}{3}\tfrac{\text{epoch}}{50000}
\)
& -0.15 \\
\hline

\textbf{Style}
& Foot clearance
& \( r_\text{fc}=
\begin{cases}
0, & \text{standing mode},\\
\sum_i \mathbf{1}[c_{f, i}](h_{f, i}-0.25)^2\|v_{f, i}\|_2, & \text{otherwise}
\end{cases}
\)
& -1.3\\

& Foot brushing penalty
& \( r_\text{fs}=\sum_i \mathbf{1}[c_{f, i}]\|v_{f, xy,i}\|_2^2 \)
& -0.5\\

& Foot airtime
& \( r_\text{air} =
\begin{cases}
\begin{aligned}
\sum_i (&\min(T_{a,i},0.25)\mathbf{1}[T_{a,i}<0.3]\\
      &+ \min(T_{s,i},0.25)\mathbf{1}[T_{s,i}<0.3])
\end{aligned}
, & \text{standing},\\
\sum_i\min(\max(T_{s,i}-T_{a,i},-0.3),0.3), & \text{otherwise}
\end{cases}
\)
& 0.1\\

& Nominal configuration
& \( r_\text{nc}=\sum_i (q_i-q_{\text{default},i})^2 / \max(\|v^{\text{cmd}}_{b,xy}\|_2^2,1) \)
& -0.005\\

& Foot collision
& \( r_\text{col}=
\mathbf{1}[\exists i:\|F_{f, i}\|_2>1 \land h_{f, i}>0.05]
\)
& -5.0 \\

& Foot contact velocity
& \( r_\text{cvel}= \sum_i v_{f, i} \,\mathbf{1}[T_{s,i}=0 \land c_{f, i}] \)
& -0.2 \\

& Roll angle limit
& \( r_\text{jl}= \mathbf{1}[|q_\text{roll}-q_{\text{default,roll}}|>\pi/4] \)
& -0.04 \\

& Body shaking
& \( r_\text{bm}=
\begin{cases}
v_{b, z}^2 + 0.5\|w_{b, xy}\|_2^2/\|v^{\text{cmd}}_{b,xy}\|_2^2, & \|v^{\text{cmd}}_{b,xy}\|_2>1,\\
v_{b, z}^2 + 0.5\|w_{b, xy}\|_2^2, & \text{otherwise}
\end{cases}
\)
& -0.4 \\

& Simultaneous contact
& \( r_\text{con}=
\begin{cases}
1, & \sum_i c_{f,i} \neq 2,\\
1, & \sum_i c_{f,i} \neq 4 \land \text{standing mode},\\
0, & \text{otherwise}
\end{cases}
\)
& -0.25 \\
\hline
\end{tabular}
\end{table*}

\begin{table*}[t]
\centering
\caption{Hyperparameters for PPO}
\vspace{1.0em}
\label{S_Tab:hyperparameters_ppo}
\renewcommand{\arraystretch}{1.20}
\begin{tabular}{ll}
\hline
\textbf{Parameter} & \textbf{Value} \\
\hline
horizon length (dt: 0.01) & 50 \\
learning rate & 3.0E-4 \\
kl threshold & 0.008 \\
discount factor & 0.99 \\
entropy coef & 0.001 \\
clip ratio & 0.2 \\
batch size & 204800 \\
mini batch size & 40960 \\
\hline
\end{tabular}
\end{table*}

\begin{table*}[t]
\centering
\caption{Student exteroceptive network parameters}
\vspace{0.5em}
\label{S_Tab:student_extero_param}
\renewcommand{\arraystretch}{1.20}
\begin{tabular}{ll}
\hline
\textbf{Parameter} & \textbf{Value} \\
\hline
CNN channels & [32, 64] \\
CNN kernel sizes & [5, 3] \\
CNN activation function & LeakyReLU \\
CNN pooling layer & MaxPool \\
CNN stride & [1, 1] \\
CNN embedding dims & 128 \\
\hline
MLP hidden size & [256, 128, 64] \\
MLP activation function & LeakyReLU \\
MLP embedding dims & 32 \\
\hline
RNN type & GRU \\
RNN layers & 1 \\
RNN hidden dims & 512 \\
RNN final activation function & tanh \\
\hline
\end{tabular}
\end{table*}


\clearpage 

\paragraph{Caption for Movie S1.}
\textbf{Robot deployment in urban campus and forest environments.}
Animated version of Figure~\ref{fig:various_scenarios}. The video demonstrates that the robot achieves agile perceptive locomotion with adaptive skills in urban campus and the forest.

\paragraph{Caption for Movie S2.}
\textbf{Auxiliary action analysis.}
Animated version of Figure~\ref{fig:dataset}C to E. Video analysis of auxiliary action during deployments.

\paragraph{Caption for Movie S3.}
\textbf{Gait selection behavior in the real world.}
Animated version of Figure~\ref{fig:trot_bound}. Video demonstrates that robot choose different gaits and motor skills for diverse obstacles in the real world.

\paragraph{Caption for Movie S4.}
\textbf{APT-RL applied to different robots.}
Auxiliary video for the Discussion section. The video demonstrates that APT-RL can be applied to different robots such as Go1, ANYmal, and KAIST HOUND in bipedal robot mode.




\clearpage


\end{document}